\documentclass[conference]{IEEEtran}
\usepackage{times}

\pdfoutput=1

\usepackage[numbers]{natbib}
\usepackage{multicol}
\usepackage[bookmarks=true]{hyperref}

\usepackage{graphics}
\usepackage{graphicx}
\usepackage{wrapfig}

\usepackage{amsfonts}
\usepackage{amssymb}

\usepackage{algorithm}
\usepackage{algorithmicx}
\usepackage[noend]{algpseudocode}
\algrenewcommand\algorithmicindent{0.75em}

\usepackage{amsmath}
\usepackage{ifthen}
\usepackage{xspace}
\usepackage{xcolor}
\let\labelindent\relax
\usepackage{enumitem}
\usepackage{amsthm}

\newcommand{\ignore}[1]{}

\def\P{\mathcal{P}}  
  
  \def\E{\mathcal{E}}
  
 \def\T{\mathcal{T}} 
 \def\V{\mathcal{V}} 
 \def\W{\mathcal{W}} 
\def\M{\mathcal{M}} \def\X{\mathcal{X}} \def\A{\mathcal{A}}

\def\SE{\mathcal{SE}}
\def\SO{\mathcal{SO}}

\newcommand{\Cpp}{C\raise.08ex\hbox{\tt ++}\xspace}

\newcommand{\NP}{{\small \ensuremath{\mathsf{NP}}\xspace}}

\newtheorem{lem}{Lemma}
\newtheorem{thm}{Theorem}
\newtheorem{thrm}{Theorem}
\newtheorem{cor}{Corollary}

\newtheorem{dft}{Definition}
\newtheorem{prob}{Problem}

\newcommand\algname[1]{\textsf{#1}\xspace}

\newcommand\astar{\algname{A*}}
\newcommand\rrt{\algname{RRT}}
\newcommand\rrtstar{\algname{RRT*}}

\newcommand\aft{\algname{AFT}}

\newcommand\rrtpara{\algname{RRT\_PARA}}
\newcommand\rcsb{\algname{RCS\_BASIC}}
\newcommand\rcsnr{\algname{RCS\_NR}}
\newcommand\rcs{\algname{RCS}}
\newcommand\rcspara{\algname{RCS\_PARA}}

\begin{document}

\title{Toward Certifiable Motion Planning for \\Medical Steerable Needles}

\author{\authorblockN{Mengyu Fu\authorrefmark{1},
Oren Salzman\authorrefmark{2},
and
Ron Alterovitz\authorrefmark{1}}
\authorblockA{\authorrefmark{1}Department of Computer Science, University of North Carolina at Chapel Hill, Chapel Hill, NC 27599, USA \\
Email: {\{mfu,ron\}@cs.unc.edu}}
\authorblockA{\authorrefmark{2}Computer Science Department, Technion - Israel Institute of Technology, Israel\\
Email: osalzman@cs.technion.ac.il}}

\maketitle

\begin{abstract}
Medical steerable needles can move along 3D curvilinear trajectories to avoid anatomical obstacles and reach clinically significant targets inside the human body.
Automating steerable needle procedures can enable physicians and patients to harness the full potential of steerable needles by maximally leveraging their steerability to safely and accurately reach targets for medical procedures such as biopsies and localized therapy delivery for cancer.
For the automation of medical procedures to be clinically accepted, it is critical from a patient care, safety, and regulatory perspective to certify the correctness and effectiveness of the motion planning algorithms involved in procedure automation.
In this paper, we take an important step toward creating a certifiable motion planner for steerable needles.
We introduce the first motion planner for steerable needles that offers a guarantee, under clinically appropriate assumptions, that it will, in finite time, compute an exact, obstacle-avoiding motion plan to a specified target, or notify the user that no such plan exists.
We present an efficient, resolution-complete motion planner for steerable needles based on a novel adaptation of multi-resolution planning.
Compared to state-of-the-art steerable needle motion planners (none of which provide any completeness guarantees), we demonstrate that our new resolution-complete motion planner computes plans faster and with a higher success rate.
\end{abstract}

\IEEEpeerreviewmaketitle

\section{Introduction}
\label{sec:intro}

Steerable needles are highly flexible medical devices able to follow 3D curvilinear trajectories inside the human body, reaching clinically significant targets while safely avoiding critical anatomical structures~\cite{Alterovitz2005_ICRA,Cowan2011_Chapter,Park2005_ICRA,Webster2006_IJRR}.
Compared with traditional rigid medical instruments, steerable needles can reduce a patient's trauma, increase safety, and provide minimally invasive access to previously inaccessible targets.
Steerable needles have been considered for a wide range of diagnostic and treatment procedures including biopsy, 
drug therapy delivery,
and radioactive seed implantation for cancer treatment~\cite{Abolhassani2007_MEP}.

\begin{figure}
    \centering
    \includegraphics[width= \linewidth]{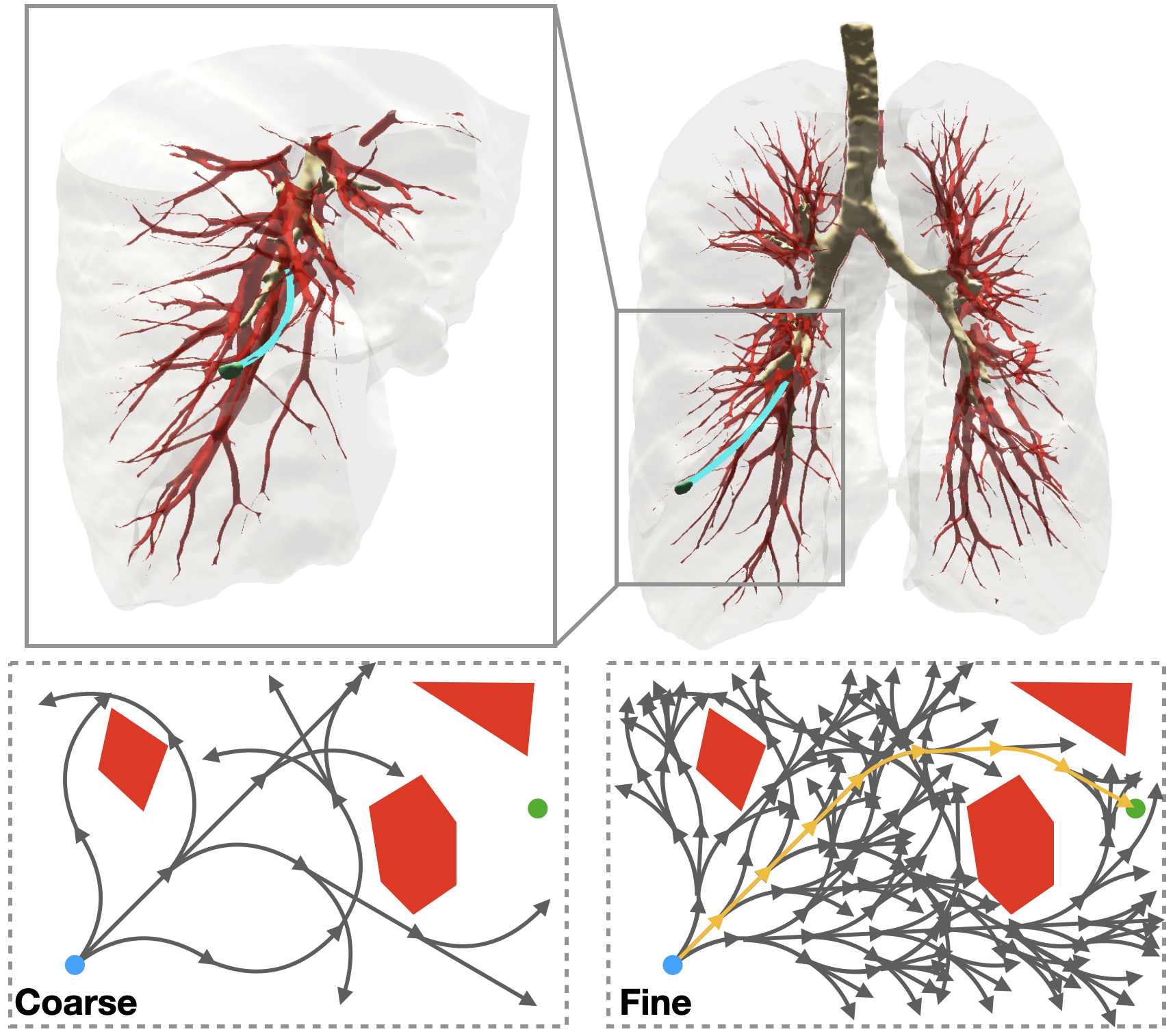}
    \caption{
    \textbf{Top:} 
    A medical steerable needle (cyan) used to reach a nodule (green) in the lung parenchyma for biopsy or cancer treatment while avoiding critical anatomical structures such as the bronchial tubes (brown) and major blood vessels (red).
    \textbf{Bottom:} 
    Our resolution-complete motion planner uses search trees built using different resolutions, illustrated here in 2D.
    A valid motion plan goes from the start configuration (blue dot) to the goal point (green dot), while avoiding obstacles (red) and satisfying kinematic constraints.
    The left search tree uses a coarse resolution and failed to find a plan while the right one uses a finer resolution and successfully generated a motion plan (yellow).
    }
    \label{fig:cover}
    \vspace{-5mm}
\end{figure}

Automating steerable needle procedures can enable physicians and patients to harness the full potential of steerable needles by maximally leveraging their steerability and ability to accurately and precisely reach targets.
Automation is critical to harnessing the full potential of these needles since the nonholonomic constraints on the needle's 3D motion coupled with the cluttered nature of anatomical environments make direct manual control unintuitive and impractical for human operators.
To automate steerable needle procedures, physicians first obtain a medical image (such as a computed tomography (CT) or magnetic resonance imaging (MRI) scan) of the relevant anatomy, from which we can segment (manually or automatically) the relevant anatomy, including the target to reach and obstacles to avoid.
The next key ingredient to the automation of steerable needle procedures is motion planning, which requires computing feasible motions to steer the needle safely around the anatomical obstacles and to the target.
An example scenario of a lung biopsy is shown in Fig.~\ref{fig:cover} (top).

For the automation of medical procedures to be clinically accepted, it is critical from a patient care, safety, and regulatory perspective to certify the correctness and effectiveness of the algorithms involved in procedure automation.
Unfortunately, \emph{no previously developed motion planner for steerable needles offers a formal guarantee that it will compute a solution, when one exists, in finite time, or notify the user that no solution exists}.
Although many steerable needle motion planners have been proposed, for all prior methods, the method either is not guaranteed to return a solution (e.g., \cite{ Duindam2010_IJRR, Favaro2018_ICRA, Hauser2009_RSS, Patil2014_TRO, Seiler2012_IJRR, Van2010_WAFR, Xu2008_ICASE}) or is not guaranteed to find a solution within a clinically reasonable distance of the target~\cite{Liu2016_RAL} when a solution exists.

As an important step toward creating a certifiable motion planner for steerable needles, we introduce the first motion planner for steerable needles that enables us to offer a guarantee under clinically appropriate assumptions that it can, in finite time, compute an exact, obstacle-avoiding motion plan to a specified target, or notify the user that no such plan exists.
In motion planning, such a guarantee is defined as \emph{completeness}~\cite{Lavalle2006_BOOK}. 
A motion planner that lacks a completeness guarantee may find solutions for only a subset of problem instances, and when no solution is found by the planner, a user has no way to distinguish whether the planner is incapable of finding an existing solution or if no solution exists. 

Providing a completeness guarantee for a steerable needle motion planner is challenging in part because motion planning for steerable needles in 3D with curvature constraints is at least \NP-hard~\cite{Kirkpatrick2011_CCCG, Solovey2020_ARXIV}. 
This challenge inspires us to consider variants of completeness relevant to medical applications. 
We note that some variants of completeness that only offer asymptotic guarantees, such as probabilistic completeness \cite{Lavalle2006_BOOK}, are not useful for needle steering since they only are guaranteed to find a solution as computation time increases to infinity, but medical applications typically require guaranteeing the planner's behaviour within a finite time.

In this paper, we focus on a specific type of completeness relevant to real-world medical applications: resolution completeness \cite{Lavalle2006_BOOK}.
Generally speaking, a resolution characterizes the discretization of some space (e.g., state space, configuration space, action space, and time).
An algorithm is resolution complete if there exists a fine-enough resolution with which the algorithm finds a plan in finite time when a qualified solution exists, and otherwise correctly returns that no such plan exists.
We illustrate at the bottom of Fig.~\ref{fig:cover} an example showing searches with different resolutions for needle steering.

In this work, we present an efficient, resolution-complete motion planner for steerable needles based on a novel adaptation of multi-resolution planning.
The planner is resolution complete, which means under some mild assumptions on the system and the solution (detailed in Sec.~\ref{sec:theory} and \ref{sec:appendix}), the planner, in finite time, is guaranteed to find a motion plan as long as the problem admits a qualified solution.
Our main contributions include:
(i)~carefully defining the motion primitives~\cite{Frazzoli2002_JGCD} used by our planner which are specifically tailored to our domain of 3D steerable needles (Sec.~\ref{subsec:motion_primitive});
(ii)~introducing a set of domain-specific optimizations that improve the efficiency of the algorithm while maintaining resolution completeness (Sec.~\ref{subsec:details});
and (iii)~providing a proof sketch to show the resolution completeness of our method (Sec.~\ref{sec:theory} and~\ref{sec:appendix}).

We demonstrate the performance of our planner in scenarios based on lung biopsy. In these scenarios, a steerable needle is deployed through a bronchoscope and must steer through the lung parenchyma (the substance of the lung outside the bronchial tubes) to a target while safely avoiding obstacles (e.g., blood vessels). We compare in simulation our planner with two existing steerable needle planners---one is sampling based while the other is search based.
Not only does our motion planner provide a resolution-completeness guarantee, but compared to prior work it also computes plans of comparable quality, faster, and with a higher success rate.
\section{Related Work}
\label{sec:related_work}

Steerable needles have many different designs, including 
bevel-tip flexible needles~\cite{Webster2006_IJRR,Cowan2011_Chapter}, 
symmetric-tip needles~\cite{Dimaio2003_TRA}, 
needles with curved stylet tips~\cite{Okazawa2005_ITM},
needles with tendon-actuated tips~\cite{Qi2014_EMBC},
and programmable bevel-tip needles~\cite{Ko2011_TRO, Secoli2016_BIOROB}.
In this paper, we focus on bevel-tip flexible needles but our approach can be easily used in any mechanical design as long as  the major kinematic constraint to consider is the curvature of the needle trajectory.

\subsection{Motion planning for steerable needles}
Early work studied planning and control for steerable needles in the 2D plane~\cite{Alterovitz2007_RSS, Asadian2011_JINT, Bernardes2012_ICRA, Reed2011_RAM}.
To fully utilize the capability of steerable needles, later work began to focus more on needle steering in 3D environments.
Duindam et al.~\cite{Duindam2010_IJRR} used inverse kinematics for planning but the planner was tested only with simple geometrically shaped obstacles and provides no theoretical guarantees.

Other planners built upon the probabilistic completeness guarantees of sampling-based methods such as the Rapidly-exploring Random Tree (\rrt)~\cite{Lavalle1998}.
Xu et al.~\cite{Xu2008_ICASE} used an \rrt variant for needle steering but showed low efficiency in computing time.
Patil et al.~\cite{Patil2014_TRO} developed an \rrt-based needle planner that guides the tree expansion by sampling in the 3D workspace (instead of the configuration space).
The efficiency obtained by sampling in the workspace and not accounting for the needle's orientation makes the planner extremely fast in practice. Unfortunately, this makes the completeness proofs of the original \rrt inapplicable and probabilistic completeness is not guaranteed.

To avoid dealing with curvature constraints directly in the \rrt algorithms, there are also hybrid methods that combine sampling and other techniques.
Favaro et al.~\cite{Favaro2018_ICRA} proposed a method that uses \rrtstar~\cite{Karaman2011_IJRR} that builds a tree embedded in the 3D workspace to generate candidate plans of low cost, followed by a smoothing step to account for the curvature constraint.
However, this decoupled approach does not provide any theoretical guarantees.

Liu et al. proposed the Adaptive Fractal Tree (\aft)~\cite{Liu2016_RAL} for needle steering and used a Graphics Processing Unit (GPU) to further speed up their algorithm.
The method uses a greedy approach for path refinement---it iteratively uses the lowest-cost path in the previous iteration for plan refinement.
However, expanding the best path of a coarse resolution does not necessarily lead to a best path of a finer resolution.
Furthermore, the authors use a cost function consisting of three factors, only one of which is the distance to the goal, also known as the targeting error.
Thus, when provided with a required targeting error, paths produced by the method are not guaranteed to adhere to this constraint since the targeting error may be sacrificed for a better cost for the other two terms.
Pinzi et al.~\cite{Pinzi2019_IJCARS} later extended \aft to account for goal orientation constraints.

Other methods focus on accounting for uncertainty during needle insertion but do not account for completeness~\cite{Hauser2009_RSS, Seiler2012_IJRR, Van2010_WAFR,Sun2015_TRO}.
To summarize, to the best of the authors' knowledge, none of the existing steerable needle planners provide provable guarantees on the planner's completeness.

\subsection{Resolution-complete motion planners}

Generally speaking, an algorithm is \textit{resolution complete} if it generates a plan to the goal whenever a solution exists at the maximal resolution and returns failure otherwise~\cite{Barraquand1991_IJRR}.
This  property guarantees that given a predefined maximal resolution, the algorithm terminates in finite time and provides a deterministic result.

Barraquand et al.~\cite{Barraquand1993_Algorithmica} proposed a planner for single/multi-body mobile robots with nonholonomic constraints.
They formally proved the planner is guaranteed to generate a solution path when the discretization of the search parameters is fine enough.
This approach was later extended by Lindemann and LaValle~\cite{Lindemann2006_ICRA} to suggest a multi-resolution approach for 2D car-like robots.
Both these works~\cite{Barraquand1993_Algorithmica, Lindemann2006_ICRA} serve as the algorithmic foundations to the planner we present in this paper.

Sampling-based planners (such as \rrt) typically ensure probabilistic completeness (i.e., such a planner is guaranteed to  find a solution, if one exists, with probability one when given infinite time). 
However, they can also be used to build resolution-complete planners given some mild assumptions on the minimal motion that the system can perform.
Cheng et al.~\cite{Cheng2002_ICRA} proposed a resolution-complete version of \rrt for systems that satisfy the Lipschitz condition.
Yershov et al.~\cite{Yershov2010_WAFR} formally analyzed the system conditions for the existence of resolution-complete planners.
Kleinbort et al.~\cite{Kleinbort2018_RAL} later analyzed the assumptions for \rrt's probabilistic completeness in kinodynamic planning. However their analysis can be adapted to resolution-completeness guarantees.

Ljungqvist et al.~\cite{Ljungqvist2017_IVS} proposed a planner for a general two-trailer system in 2D.
They used a two-point boundary value problem (2pBVP) solver to generate a set of motion primitives connecting 2D grid points.
Their planner is resolution-complete and resolution-optimal with respect to the resolution in the configuration space,
which means the planner generates a plan with minimal cost among all solutions that can be represented as a sequence of motion primitives.
Most of the above-mentioned planners can be used to plan for 2D nonholonomic robots.
However, none account explicitly for the challenges of planning with curvature constraints in 3D, where the dimension of the search space is higher and there is no efficient 2pBVP solver.
In this work, we provide a planner for 3D needle steering that is both efficient in practice and is guaranteed to be resolution complete.
\section{Problem Definition}
\label{sec:pdef}

In this work, we consider  steerable needles that operate in a 3D workspace~$\W \in \mathbb{R}^3$, which is cluttered with obstacles~$\W_{\rm obs} \subset \W$.
We define the configuration space (or C-space) of the steerable needle as $\X \subset \SE(3)$.
Each configuration~$\mathbf{x} = (p, q) \in \X$ uniquely defines the pose (i.e., position~$p \in \mathbb{R}^3$ and orientation~$q \in \SO(3)$) of the needle tip.
We define a projection function ${\rm Proj}(\cdot) : \X \rightarrow \W$ that projects configurations to points in the workspace, i.e., ${\rm Proj}(\mathbf{x}) = p$.
A configuration~$\mathbf{x}$ is \textit{collision free} if ${\rm Proj}(\mathbf{x}) \notin \W_{\rm obs}$, and is \textit{in collision} otherwise.
The union of all collision-free configurations is denoted as~$\X_{\rm free}$.
Since we assume the needle shaft perfectly follows its tip, a motion plan of the needle can be uniquely defined as a trajectory~$\sigma: [0, 1] \rightarrow \X$.
And such a motion plan~$\sigma$ is \textit{collision free} if all configurations along the trajectory are collision free. Namely, $\forall s \in [0, 1], \sigma(s) \in \X_{\rm free}$.

\begin{figure}
    \centering
    \includegraphics[width=0.6\linewidth]{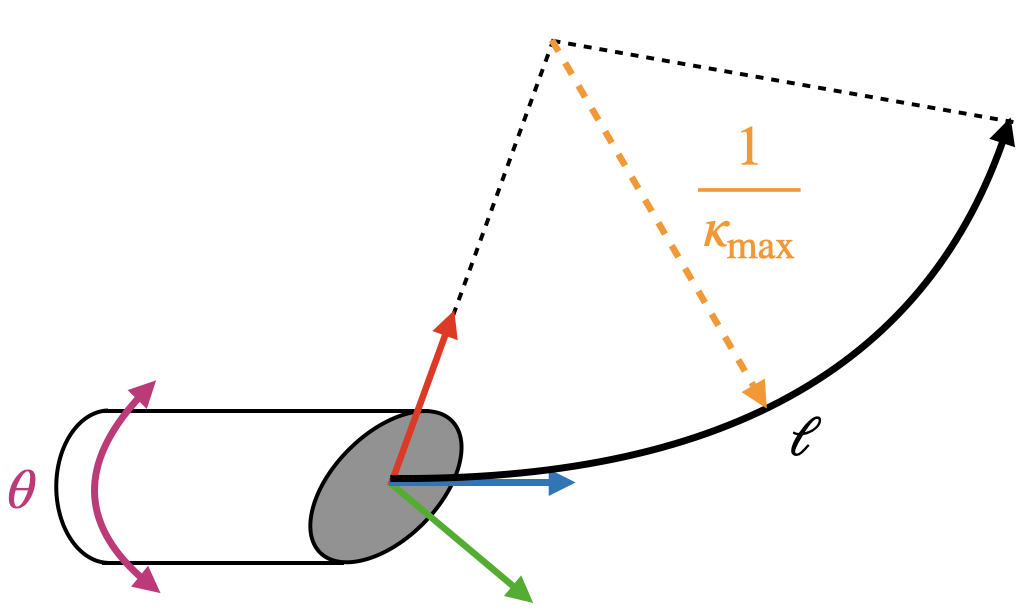}
    \caption{The kinematics of a bevel-tip steerable needle.
    The needle can be inserted (characterized by~$\ell$) and axially rotated at its base (characterized by~$\theta$).}
    \label{fig:needle-control}
\vspace{-5mm}
\end{figure}

We also need to consider the kinematics of the steerable needle.
We specifically consider steerable needles that are highly flexible and have an asymmetric tip (e.g., a bevel) \cite{Alterovitz2005_ICRA,Cowan2011_Chapter,Park2005_ICRA,Webster2006_IJRR}; the asymmetric tip exerts asymmetric forces on the tissue in front of the needle tip, and the  high flexibility enables the needle to curve substantially at maximum curvature~$\kappa_{\max}$ as it moves through the tissue.
Furthermore, rotating the needle axially at its base changes the direction of the needle's asymmetric tip, enabling the needle to change its direction of steering.
See Fig.~\ref{fig:needle-control} for an illustration.

We say a motion plan is (kinematically) \textit{feasible} if it never exceeds the maximum curvature~$\kappa_{\max}$.
A \textit{valid} motion plan for the needle is both collision free and feasible.
We also assume there exists a resolution describing the smallest interval or precision of the achievable motions,
which may be limited by the physical system's hardware (e.g., motor, encoders, controller, etc.) and its interaction with the environment.
In this paper, we determine this finest resolution by considering the hardware's ability to measurably change the steerable needle tip's position and orientation in tissue.
Considering real-world effects such as torsional wind up of the needle shaft during actuation, the control resolution of the needle tip is coarser than the control resolution of the needle base where motors directly apply controls.
Thus, we are not using minimal motions of the motors.
Instead, we consider the minimal motions the tip of the needle can perform.
We assume there exists a lower-level controller taking care of controlling the tip to the desired pose, as is common in needle steering systems.

We are now ready to state the steerable needle motion planning problem.

\begin{prob}
\label{prob:MP}
A steerable needle motion planning problem is defined as the tuple
$\Delta = (\X, \W_{\rm obs}, \mathbf{x}_{\rm start}, p_{\rm goal}, \tau, \ell_{\max}, \kappa_{\max})$,
where~$\W_{\rm obs}$ is the obstacle set,
$\mathbf{x}_{\rm start}$ is the start configuration,
$p_{\rm goal} \in \W$ is the goal point,
$\tau > 0 $ is the goal tolerance,
$\ell_{\max}$ is the maximum insertion length,
and $\kappa_{\max}$ is the maximum curvature.
The problem calls for computing a valid motion plan~$\sigma$ that satisfies:
(i)~$\sigma(0) = \mathbf{x}_{\rm start}$, 
(ii)~the Euclidean norm $\|{\rm Proj}(\sigma(1)) - p_{\rm goal}\|_2 \leq \tau$,
and
(iii)~trajectory length $\ell(\sigma) \leq \ell_{\max}$.
\end{prob}

As we show in our later discussion (in Sec.~\ref{sec:theory} and \ref{sec:appendix}), for any given instance of Problem~\ref{prob:MP}, under some mild assumptions, there exists some fine-enough resolution~$R_{\min} = \{\delta\ell_{\min}, \delta\theta_{\min}\}$ 
(corresponding to the needle's insertion and axial rotation, respectively)
for which our planner is guaranteed to find a solution in finite time (when one exists) or to indicate that no solution exists.

\section{Method}
\label{sec:method}

\subsection{Overview}

Our needle planner builds a search tree $\T = (\V, \E)$ embedded in the  C-space with~$\mathbf{x}_{\rm start}$ as its root.
Each node~$v \in \V$ is associated with a configuration~$\mathbf{x}_v \in \X$, and each edge $e = (v, u) \in \E$ represents the transition from $\mathbf{x}_v$ to $\mathbf{x}_u$.
To expand a node~$v \in \V$, we construct new nodes (children of~$v$) with \textit{motion primitives} (to be explained shortly in Sec.~\ref{subsec:motion_primitive}), which are pre-defined feasible motions.
A child node~$v_{\rm child}$ is accepted and added to the search tree if the trajectory from $v$ to~$v_{\rm child}$ is collision-free and $v_{\rm child}$ is valid (will be detailed in Sec.~\ref{subsec:algorithm}).
The search tree grows until there is some node~$v$ with configuration~$\mathbf{x}_v$ whose projection is inside the $\tau$-neighborhood of $p_{\rm goal}$ (condition (ii) in Problem~\ref{prob:MP}).

A key aspect of our search method (which is similar in nature to other search-based planners~\cite{Lindemann2006_ICRA}) is to use a set of motion primitives defined using \emph{multiple} resolutions.
Instead of expanding each node in our search tree using the entire set of motion primitives, we start with coarse motion primitives and use finer motion primitives as the search progresses.
Thus, we start (Sec.~\ref{subsec:motion_primitive}) by describing the parameters required to define a motion primitive.
After that, we continue (Sec.~\ref{subsec:multi_resolution}) to detail a hierarchy of motion primitives together with an ordering that will be used in our search algorithm.
We then describe our search algorithm in detail (Sec.~\ref{subsec:algorithm}) and elaborate on the method we use to handle ``similar'' states, also known as \emph{duplicate detection}~\cite{Du2019_IROS} (Sec.~\ref{subsec:duplicate_detection}).
We conclude this section with some implementation details (Sec.~\ref{subsec:details}).

\subsection{Motion Primitives}
\label{subsec:motion_primitive}

\begin{figure}
    \centering
    \includegraphics[width=\linewidth]{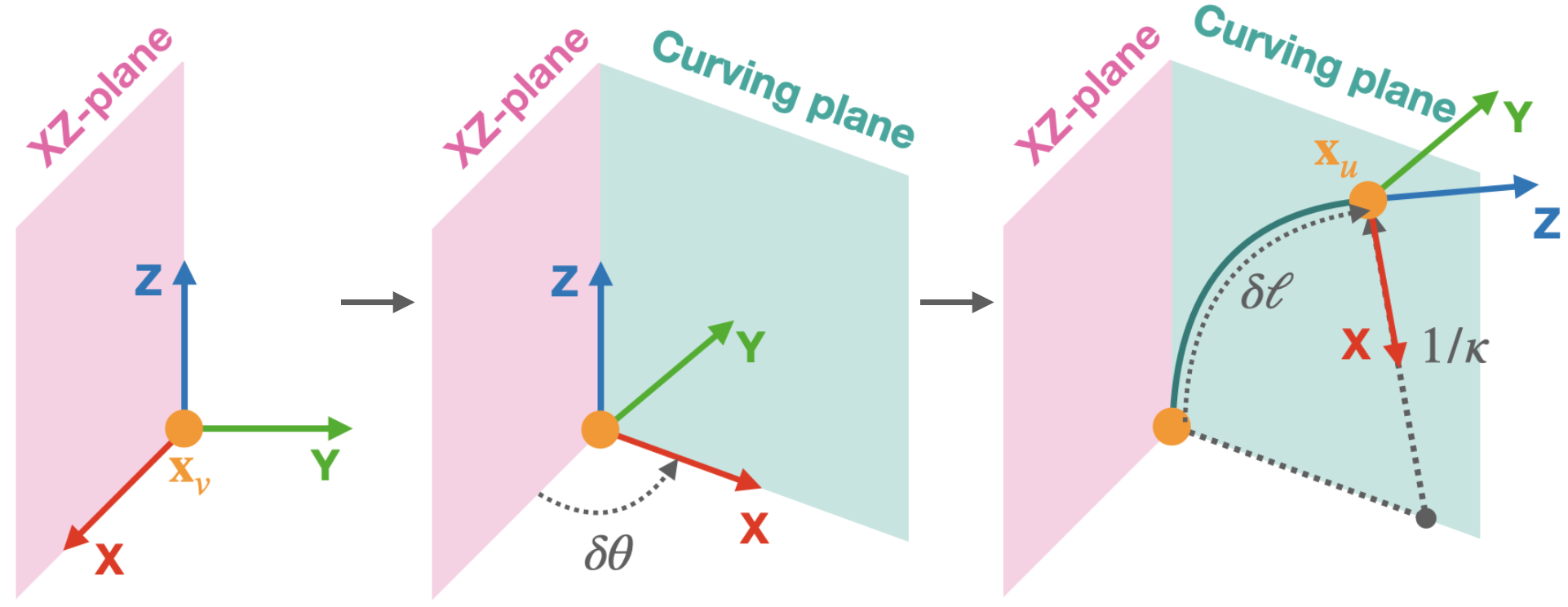}
    \caption{A motion primitive is a circular arc defined as $\M = (\kappa, \delta\ell, \delta\theta)$.
    The circular arc (dark green) lies in the curving plane (light green) that contains the Z-axis (blue) at the start configuration~$\mathbf{x}_v$.
    $\kappa$ is the curvature of the arc,~$\delta\theta$ is the angle between the curving plane and the XZ-plane, and $\delta\ell$ is the length of the arc.
    The figures show step-by-step how the child configuration $\mathbf{x}_u = \mathbf{x}_v \oplus \M$ is generated.
    }
    \label{fig:primitive}

\vspace{-5mm}
\end{figure}

Motion primitives, introduced by Frazzoli et al.~\cite{Frazzoli2002_JGCD}, have been used in many motion planners~\cite{Islam2019_ICAPS, Islam2020_ICRA,Lindemann2006_ICRA, Pivtoraiko2011_IROS,Ljungqvist2017_IVS}.
In our setting, the motion primitives are a set of predefined kinematically feasible local motions.
Roughly speaking, a motion primitive defines
with what curvature the needle curves, how far the needle steers, and in what direction (see Fig.~\ref{fig:primitive}).
Since for each motion primitive, the curvature~$\kappa$ is explicitly defined, a motion primitive is guaranteed to be kinematically feasible as long as $\kappa \leq \kappa_{\max}$.
As we will see in the proofs (\ref{sec:appendix}), our definition of motion primitives guarantees resolution completeness, and the experiments show that the definition also the enables computation efficiency of our algorithm.

More formally, 
to steer from configuration $\mathbf{x}_v$, 
a motion primitive is defined as a 
three-tuple~$\M = (\kappa, \delta\ell, \delta\theta)$, 
where~$\kappa \in [0, \kappa_{\max}]$ is the curvature, 
$\delta\ell > 0$ is the length of the circular arc,
and $\delta\theta \in [0, 2\pi)$ is the angle between the curving plane and the XZ-plane of $\mathbf{x}_v$ (see Fig.~\ref{fig:primitive}).
Thus the \textit{action space} (or motion space) can be defined as $\A \subset \mathbb{R}^3$, which is the set of all motion primitives.
We use $\mathbf{x}_u = \mathbf{x}_v \oplus \M$ to denote the operation of extending~$\mathbf{x}_v$ with motion primitive~$\M$ and obtaining the resultant configuration~$\mathbf{x}_u$.
See Fig.~\ref{fig:primitive} for a step-by-step determination of $\mathbf{x}_u$.
In the context of a search tree, by a slight abuse of notation, $u = v \oplus \M$ denotes the resultant node~$u$, obtained by extending node $v$ with the motion primitive~$\M$.
We call $\M$ the \textit{extending primitive} of node $u$.

Using motion primitives allows pre-computing intermediate configurations and thus saving computation efforts during planning by transforming these configurations to the frame defined by~$\mathbf{x}_v$.
Since the trajectory produced with one motion primitive is a circular arc, it is possible to densely interpolate the trajectory for collision-checking purposes.

In the following sections, we show that $\delta\ell$ and $\delta\theta$ are gradually refined in the algorithm.
In contrast, we keep a fixed set of curvatures, $\{0, \kappa_{\max}\}$, for all motion primitives.
As we will see (Sec.~\ref{sec:theory} and \ref{sec:appendix}) this does not hinder the guarantees provided by our approach.
Moreover, as we demonstrate in our experiments (Sec.~\ref{sec:results}), these primitives, coupled with our planner allow us to efficiently compute paths for non-trivial instances where other planners fail.

\subsection{Motion Primitive Hierarchy}
\label{subsec:multi_resolution}

\begin{figure}
    \centering
    \includegraphics[width=\linewidth]{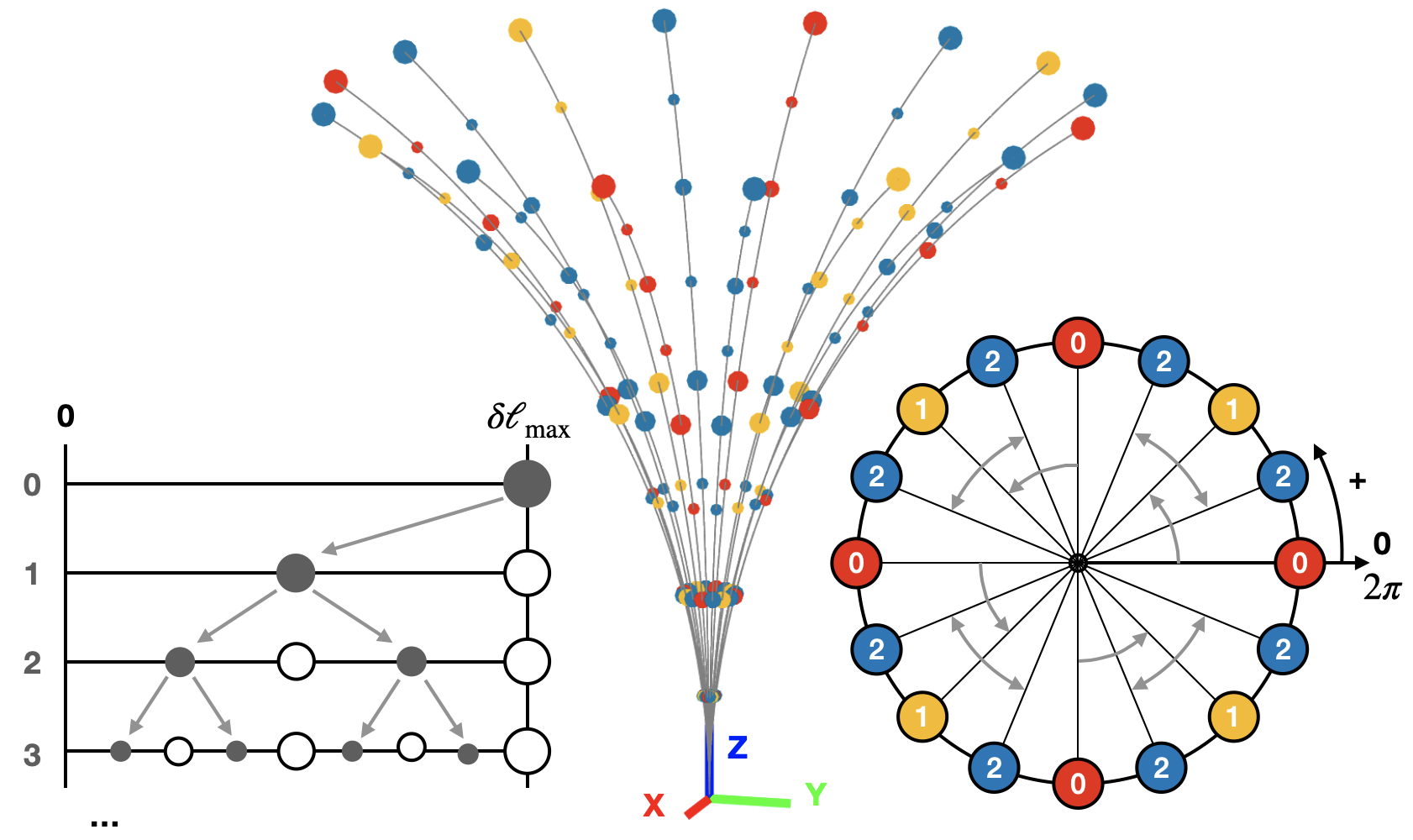}
    \caption{
    \textbf{Visualization of length and angle levels.}
    \textbf{Left:} 
        Visualization of length levels.
        Smaller node sizes correspond to higher length levels.
        The first length level ($l_{\ell} = 0$) corresponds to motion primitives of maximal length~($\delta \ell_{\max}$).
        As $l_{\ell}$ increases, the resolution of length becomes higher.
        The gray arrows illustrate how motion primitives with the first 4 length levels are generated during refinement.
    \textbf{Right:} 
        Visualization of angle levels.
        Nodes with angle levels 0, 1, 2 are shown in red, yellow, and blue, respectively.
        The first angle level ($l_{\theta} = 0$) corresponds to motion primitives of $\delta\theta = \{0, \frac{\pi}{2}, \pi, \frac{3\pi}{2}\}$.
        As $l_{\theta}$ increases, the resolution of orientation becomes higher.
        The circular arrows illustrate how nodes with the first three angle levels are generated during refinement.
    \textbf{Middle:}
        3D visualization of length and angle levels.
    }
    \label{fig:levels}

\vspace{-5mm}
\end{figure}

Our algorithm uses a sequence of motion primitives, whose resolution changes from coarse to fine.
The coarsest motion primitives are defined by some parameters $\delta\ell_{\max}$ and~$\delta\theta_{\max}$.
In our implementation and examples (e.g., Fig.~\ref{fig:levels}) we have that $\delta\theta_{\max}=\frac{\pi}{2}$ and $\delta\ell_{\max} > 0$ is a user-given parameter.

Since $\delta\theta \in [0, 2\pi)$ and $\delta\theta_{\max}=\frac{\pi}{2}$, there exist four orientations ($\delta\theta \in \{ 0, 0.5\pi, \pi, 1.5\pi \}$) that have the coarsest orientation (see Fig.~\ref{fig:levels}).
There exists only one coarsest length, which is $\delta\ell_{\max}$, since path length is accumulated when we expand a node.
To characterize how fine the resolution of a motion primitive~$\M = (\kappa, \delta\ell, \delta\theta)$ is, we define the notions of length level~$l_{\ell}$ and angle level~$l_{\theta}$. 
More formally, 
\begin{equation*}
\begin{split}
    l_{\ell}(\M) &= \min\{l \in \mathbb{Z} ~\vert~ l \geq 0, {\rm MOD}(\delta\ell, 2^{-l}\cdot\delta\ell_{\max}) = 0\}, \\
    l_{\theta}(\M) &= \min\{l \in \mathbb{Z} ~\vert~ l \geq 0, {\rm MOD}(\delta\theta, 2^{-l}\cdot\delta\theta_{\max}) = 0\},
\end{split}
\end{equation*}
where ${\rm MOD}(\cdot)$ is the modulo operation.

For a motion primitive~$\M = (\kappa, \delta\ell, \delta\theta)$, we refine the resolution of  both  the insertion $\delta\ell$ and  the orientation~$\delta\theta$.
The new motion primitives constructed by refining $\delta\ell$ are:
\begin{equation}
\label{eq:refine-ell}
    \M_{\ell\pm} = (\kappa, \delta\ell \pm 2^{-(l_{\ell}(\M)+1)}\cdot\delta\ell_{\max}, \delta\theta).
\end{equation}
Similarly, 
the motion primitives constructed by refining~$\delta\theta$ are:
\begin{equation}
\label{eq:refine-theta}
    \M_{\theta\pm} = (\kappa, \delta\ell, \delta\theta \pm 2^{-(l_{\theta}(\M)+1)}\cdot\delta\theta_{\max}).
\end{equation}
It is straight-forward to see that the 
refined motion primitives~$\M_{\ell-}$ and~$\M_{\ell+}$ both have a length level of $l_{\ell}(\M)+1$
and the 
refined motion primitives~$\M_{\theta-}$ and~$\M_{\theta+}$ both have an angle level of $l_{\theta}(\M)+1$ (see Fig.~\ref{fig:levels}).

Note that 
when refining a motion primitive with $l_{\ell}(\M) = 0$ (resp. $l_{\theta}(\M) = 0$), we ignore $\M_{\ell+}$ (resp. $\M_{\theta-}$) as they both exceed the range of exploration.
 
\begin{figure}
    \centering
    \includegraphics[width=0.95\linewidth]{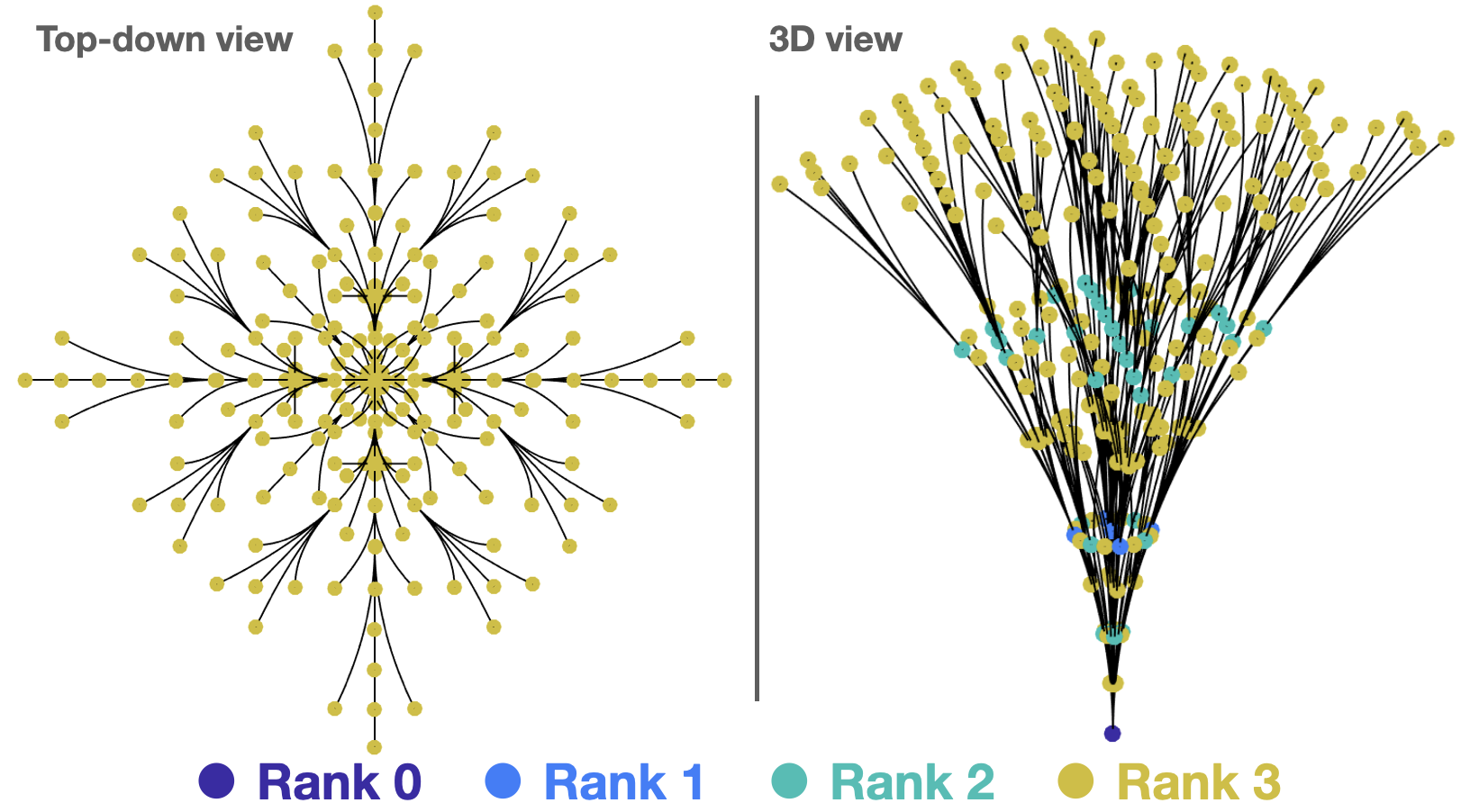}
    \caption{
    Nodes of the first four ranks.
    We use motion primitives with $\kappa = 0$ (straight lines) and $\kappa = \kappa_{\max}$ (arcs with maximum curvature).
    }
    \label{fig:node_ranks}
\end{figure}

Similar to Lindemann and LaValle~\cite{Lindemann2006_ICRA}, our search algorithm expands nodes according to a node's \textit{rank}.
Rank captures both the depth of a node in the search tree and the fineness of resolution along the branch connecting the node from the root.
We define the rank of the root node to be zero, the rank of any other node~$v$ is recursively defined as:
\begin{equation}
    {\rm Rank}(v) = {\rm Rank}(v.{\rm parent}) + l_{\ell}(\M_v) + l_{\theta}(\M_v) + 1.
\label{eq:rank}
\end{equation}
For a visualization, see Figs.~\ref{fig:levels} and~\ref{fig:node_ranks}.

\subsection{Algorithm Description}
\label{subsec:algorithm}
\begin{algorithm}[t!]
\caption{MultiResolutionSearch\\
         \textbf{Input:} $\W_{\rm obs}, 
                          \mathbf{x}_{\rm start}, 
                          p_{\rm goal}, 
                          \tau,
                          \kappa_{\max},
                          \ell_{\max}, 
                          \delta\ell_{\max}$}
    \begin{algorithmic}[1]
        
        \State{$\Theta \leftarrow \{0, \frac{\pi}{2}, \pi, \frac{3\pi}{2}\}, K \leftarrow \{0, \kappa_{\max}\}$}
        \label{line:init_coarse}
        \State{root $\leftarrow (\mathbf{x}_{\rm start}, 0)$}
        \Comment{{\footnotesize The root has rank 0}}
        \State{OPEN $\leftarrow \{$root$\}$, CLOSED $\leftarrow \emptyset$}
        \label{line:init_sets}
\vspace{1mm}
        \While{not OPEN.empty()}
            \State{$v \leftarrow$ OPEN.extract()}
            \label{line:extract}
\vspace{1mm}
            \If{Valid($v, \W_{\rm obs}, p_{\rm goal}, \ell_{\max}$)}
            \label{line:valid}
                \If{\textbf{not} existSimilarConfig($v$, CLOSED)}
                \label{line:similar_node}
                    \If{Terminate($v, p_{\rm goal}, \tau$)}
                        \State{\textbf{return} RetrievePlan($v$)}
                    \EndIf
\vspace{1mm}
                    \For{$\M \in$ Primitives($K, \delta\ell_{\max}, \Theta$)}
                    \label{line:expand}
                        \State{OPEN.insert($v \oplus \M$)}
                        \label{line:expand_1}
                    \EndFor
                    \State{CLOSED.insert($v$)}
                    \label{line:add_to_closed}
                \EndIf
            \EndIf
\vspace{1mm} 
            \If{$v$ != root}
                \For{$\M \in$ RefinedPrimitives($\M_v$)}
                \label{line:refine-primitive}
                    \State{OPEN.insert($v.{\rm parent} \oplus \M$)}
                    \label{line:finer_nodes}
                \EndFor
            \EndIf
        \EndWhile
\vspace{1mm}
        \State{\textbf{return} NULL}
    \end{algorithmic}
\label{alg:main}
\end{algorithm}

We run an \astar-like search where nodes are ordered according to their rank (Eq.~\ref{eq:rank}).
A distinctive feature from (vanilla) \astar is that when we expand a node, we also increase the resolution of the motion primitives used to expand its parent and add nodes using these finer motion primitives to the search's priority queue.
The rest of this section formalizes this idea.

Alg.~\ref{alg:main} shows the pseudocode of our needle planner.
We first initialize the coarsest orientations and the curvature set (line~\ref{line:init_coarse}), then initialize the OPEN list and CLOSED set (line~\ref{line:init_sets}).
The search algorithm then iteratively extracts nodes from the OPEN list (line~\ref{line:extract}), where nodes are ordered in a monotonically non-decreasing order according to their rank.

Only at this point (line~\ref{line:valid}) the extracted node is validated (also known as \emph{lazy}  validation~\cite{Hauser2015_ICRA,Mandalika2019_ICAPS}).
Validation of node $v$ involves ensuring that:
(i)~the accumulated trajectory length should not exceed the maximum insertion length $\ell_{\max}$;
(ii)~the goal point should be inside or close to the reachable region of $\mathbf{x}_v$ (Sec.~\ref{subsec:details});
(iii)~$v$ should not be a duplicated node (Sec.~\ref{subsec:details});
and that
(iv)~the circular arc connecting $v.{\rm parent}$ and $v$ should be collision-free.
An invalid node will be rejected and discarded.
For a valid node~$v$, we further check if there exists any similar configuration in the CLOSED set in order to avoid considering highly similar configurations (Sec.~\ref{subsec:duplicate_detection} and \ref{sec:appendix}).
A valid node without a similar configuration is accepted, expanded, and added to the CLOSED set (lines~\ref{line:expand}-\ref{line:add_to_closed}).
The search terminates if the associated configuration of the accepted node satisfies the goal tolerance.

In our search algorithm, only the coarsest child nodes are added to the OPEN list during the initial expansion of a node (lines~\ref{line:expand}-\ref{line:expand_1}).
But additional child nodes, created with finer motion primitives, are added when the coarse child nodes are extracted from the OPEN list (line~\ref{line:finer_nodes}).
More specifically, when node~$v$ is extracted, we refine its extending motion primitive~$\M_v$ following Eq.~\ref{eq:refine-ell} and~\ref{eq:refine-theta} (line~\ref{line:refine-primitive}), and use the refined motion primitives~$\M_{\ell\pm}$ and $\M_{\theta\pm}$ to expand $v.{\rm parent}$.

\subsection{Cutoff Resolution}
\label{subsec:cutoff-resolution}

As specified in Sec.~\ref{sec:pdef}, for a physical needle-steering robot there exists some smallest interval or precision of the achievable motions, which induces the minimal insertion and axial rotation $\delta\ell_{\min}$ and $\delta\theta_{\min}$, respectively.
We term $\delta\ell_{\min}$ and $\delta\theta_{\min}$ as the \textit{cutoff resolution} and stop adding refined nodes when the extending motion primitive~$\M$ satisfies
$2^{-l_{\ell}(\M)}\cdot\delta\ell_{\max} < \delta\ell_{\min}$
or
$2^{-l_{\theta}(\M)}\cdot\delta\theta_{\max} < \delta\theta_{\min}$.

\subsection{Duplicate Detection}
\label{subsec:duplicate_detection}

To avoid re-expanding the same or highly similar nodes multiple times, search-based planners often employ \emph{duplicate detection}~\cite{Du2019_IROS} that prunes so-called ``duplicate'' nodes.
To prune duplicate nodes and enable the planner to rapidly explore the entire C-space, we reject a node if there already exists a similar configuration in the search tree (line~\ref{line:similar_node}).
More formally, we reject node $v$ with configuration~$\mathbf{x}_v$ if 
$\exists u \in \V, \rho(\mathbf{x}_u, \mathbf{x}_v) < d_{\rm sim}$, where $d_{\rm sim} > 0$ is a radius we use to identify similar configurations.
Here, $\rho(\cdot)$ is a distance metric defined on $\X$ which in our work is defined as 
\begin{equation}
\label{eq:metric}
    \rho(\mathbf{x}_u, \mathbf{x}_v) = \|p_u - p_v\|_2 + \alpha\cdot{\rm dist}_\sphericalangle(q_u, q_v),
\end{equation}
where $\alpha > 0$ is a weighting parameter and ${\rm dist}_\sphericalangle()$ is the angular distance between two orientations.
Note that to guarantee resolution completeness, 
the value of
~$d_{\rm sim}$ depends on other system parameters detailed in Sec.~\ref{sec:theory} and \ref{sec:appendix}.

\subsection{Implementation Details}
\label{subsec:details}

We now describe several implementation details used to further speed up our approach.
To distinguish between different implementations of our approach we refer to
the basic version of our Resolution-Complete Search (i.e., without the implementation details described below) as \rcsb
and to the (basic) version that does not use similar-node rejection (i.e., when not performing the test in line~\ref{line:similar_node}) as \rcsnr.
The versions that use all the following implementation details without and with parallelization (explained shortly) will  be referred to as  \rcs and \rcspara, respectively.

\subsubsection{Early pruning by testing for goal reachability}
\begin{figure}
    \centering
    \includegraphics[width=0.85\linewidth]{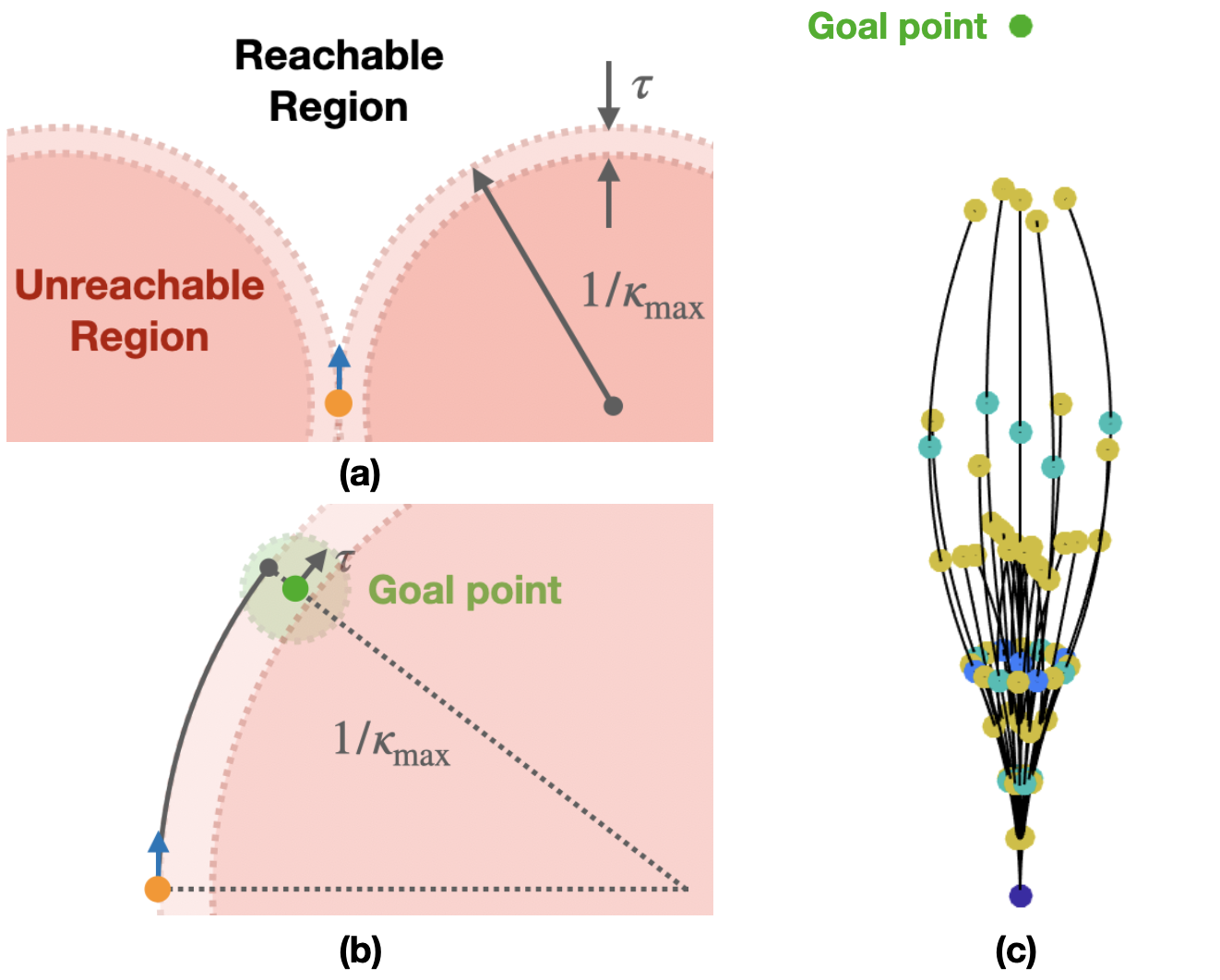}
    \caption{
    \textbf{(a)} An illustration of reachable and unreachable regions in 2D.
    The case in 3D is similar. The unreachable region can be generated by rotating the circles around the Z-axis (blue vector), which creates a donut-like shape in 3D that is unreachable.
    It also visualizes how we check goal-reachability when considering tolerance $\tau$.
    We reject a configuration if the relative position of $p_{\rm goal}$ falls in the inner region (darker orange).
    \textbf{(b)} The algorithm creates a direct connection to the goal when $p_{\rm goal}$ is outside but still close to the boundary of the reachable region.
    We use a circular arc with curvature $\kappa_{\max}$ to steer towards $p_{\rm goal}$ and the arc stops at the closest point to $p_{\rm goal}$.
    \textbf{(c)} An example of valid nodes with rank 0-3 after checking goal reachability.
    }
    \label{fig:reachable}

\vspace{-5mm}
\end{figure}

We can prune away nodes that, due to curvature constraints,  cannot be part of a path that reaches the goal (see Fig.~\ref{fig:reachable} for a 2D illustration).
The curvature constraint defines so-called ``unreachable regions'' of a node and testing if the goal~$p_{\rm goal}$  belongs to a node's unreachable  region can be done efficiently (see Fig.~\ref{fig:reachable}).
Such nodes are pruned away and not expanded.

However, recall that we allow some goal tolerance $\tau$. 
Thus, instead of requiring the goal point to be inside a node's reachable region, we only require that the distance between~$p_{\rm goal}$ and the boundary of the reachable region is smaller than $\tau$.

Our model allows a needle to make ``U-turns'' and reach the region we currently mark as unreachable.
But in our specific setting, the needle tip cannot (physically) turn more than $90^{\circ}$ as the needle might buckle and shear through the tissue, so we discard such motions.
Thus we don't need to account for a needle entering the unreachable region due to a ``U-turn''.

\subsubsection{Direct goal connection}

For each node~$v$ that is added to the search tree  with corresponding configuration~$\mathbf{x}_v$,  we attempt to connect~$\mathbf{x}_v$ to the goal point~$p_{\rm goal}$  with a circular arc (a similar technique is used in the \rrt-based needle planner~\cite{Patil2014_TRO}).
This arc lies in the plane that is determined by the tangent vector of~$\mathbf{x}_v$ and~$p_{\rm goal}$, and its  curvature can be computed according to the relative position of~$\mathbf{x}_v$ and~$p_{\rm goal}$.

If $p_{\rm goal}$ lies outside the reachable region of~$\mathbf{x}_v$ but the distance between $p_{\rm goal}$ and the boundary of the reachable region is no larger than $\tau$, we steer the needle in the plane following a circular arc of curvature $\kappa_{\max}$ to the point closest to $p_{\rm goal}$.
When the circular arc is collision-free, a solution has been found and we terminate the search.
This approach can often dramatically speed up the search.

\subsubsection{Equivalent node pruning}
As we use a multi-resolution approach, there may exist multiple nodes  representing the exact same configurations.
Our approach for rejecting similar nodes (Sec.~\ref{subsec:duplicate_detection}) can be used to reject equivalent ones.
However, testing if two nodes are equivalent is more efficient and saves future computationally expensive collision checking.

As we are refining the arc length~$\delta\ell$ and orientation~$\delta\theta$ simultaneously, it is possible for a node to be expanded more than once with the same motion primitive: first as a node with finer arc length, then as a node with finer orientation.
To avoid extending the same node with the same motion primitive, 
we give each motion primitive a unique index and 
the parent node keeps record of which motion primitives have been explored and allows only unexplored motion primitives when adding finer nodes (line~\ref{line:finer_nodes}).

\subsubsection{Parallelism}
One of the most time-consuming tasks in our search algorithm is processing a node after it is extracted from the OPEN list (namely, evaluating if the path to this node is collision free, computing the relevant motion primitives for its parent node and the corresponding new nodes).
To this end, we implemented a multi-threaded version of the algorithm where each thread is tasked with processing a node extracted from the OPEN list.
This enables processing nodes in parallel while maintaining the correctness of the algorithm by adding standard locking mechanisms to the shared data structures (i.e., OPEN list and CLOSED set).
\section{Theoretical Guarantees}
\label{sec:theory}

In this section we state and give a proof overview of the theoretical guarantees that our algorithm provides.
We start with some general definitions pertaining to the notion of resolution completeness adapted from LaValle~\cite{Lavalle2006_BOOK}.
Unfortunately, their generality requires masking important problem-related details such as, ``is planning defined in the C-space or in the control space?'' or ``what are the specific assumptions on the system?''
This is also the reason that existing proofs (e.g.,~~\cite[Appendix A]{Barraquand1991_IJRR} and~\cite[Thm. 5.2]{Cheng2002_ICRA}) cannot be used as is.
Thus, we quickly move to the specific setting of motion planning for steerable needles which requires  specifying the exact problem-related details and definitions.
Here we start with an overview of our proof explaining where we rely on the aforementioned proofs and where we are required to account for our specific domain and planner.

\subsection{General resolution-related definitions}
\begin{dft}[Resolution]
\label{dft:resolution}
    Resolution is a finite set of parameters 
    $R = \{r_0, r_1, ..., r_n\}$, 
    where each $r_i \in R$ characterizes the discretization of some space (e.g. state space, configuration space, action space, and time), and the smaller $r_i$ is, the finer the corresponding resolution is.
\end{dft}

\begin{dft}[Resolution completeness]
\label{dft:resolution-complete}
    For a general motion planning problem $\Delta$, a planner $\P$ is \textit{resolution complete} if when a so-called qualified solution to $\Delta$ exists, there exists some resolution $R_{\min}$ such that running $\P$ with resolution $R_{\min}$ on~$\Delta$ finds a solution in finite time.
\end{dft} 

Clearly the above definition is more a general intuition than a precise definition.
We need to define what a ``qualified solution is'' and what ``running $\P$ with resolution~$R_{\min}$ on~$\Delta$'' means.
These notions together with our main theoretic result (Thm.~\ref{thm:resolution_complete2}) are formalized in \ref{sec:appendix}.

\subsection{Proof overview}

As a first step we need to state how Def.~\ref{dft:resolution} is instantiated in our setting.
Here, the resolution is a pair 
$R = \{r_{\ell}, r_{\theta}\}$ that characterizes the action space (namely, the insertion~$\delta\ell$ and rotation~$\delta\theta$ of the needle).
However, this geometric characterization of the needle motion is a simplification of the way we control a needle in practice---via insertion and rotation \emph{velocity}.
This difference is important as the relative insertion and rotation velocity creates paths that have curvatures ranging between zero and $\kappa_{\max}$.
In contrast, our motion primitives either follow a straight line or a path of maximal curvature~$\kappa_{\max}$.
Thus, the first part of our proof shows that considering these two fixed curvatures allows us to approximate any path arbitrarily well.
This is done using the notion of \emph{duty cycling}~\cite{Minhas2007_EMBC} and is detailed in Sec.~\ref{subsec:duty-cycling} of \ref{sec:appendix}.
The original idea in~\cite{Minhas2007_EMBC} is designed specifically for bevel-tip needles.
We look at the problem from a geometric perspective  and decouple the guarantees and the needle mechanism, thus making it valid for needles with different designs.

The second part of our proof, detailed in Sec.~\ref{subsec:fixed-length-primitives} of \ref{sec:appendix}, states that any path that adheres to some mild assumptions can be approximated arbitrarily well by a very specific set of motion primitives---those with some \emph{fixed} resolution.
This is a somewhat technical but important step---it will allow us to argue that as long as the cutoff resolution (defined in Sec.~\ref{subsec:cutoff-resolution}) is fine enough, 
our algorithm \rcsnr is guaranteed to find a solution in finite time (when no node rejection is applied).
Here we adapt the original proof by Barraquand et al.~\cite[Appendix A]{Barraquand1991_IJRR} that considers paths in a two-dimensional workspace.
As our needle moves in a 3D workspace, we cannot use the proof as-is and detail some required  adaptations.

These parts are summarized in the following theorem (stated informally to avoid using notations defined in~\ref{sec:appendix}).
\begin{thrm}[Resolution completeness of \rcsnr]
\label{thrm:resolution_complete1}
    Let $\Delta = (\X, \W_{\rm obs}, \mathbf{x}_{\rm start}, p_{\rm goal}, \tau, \ell_{\max}, \kappa_{\max})$ be a steerable needle motion planning problem.
    Under the assumption that the system is Lipschitz continuous and there exists a traceable solution with non-zero clearance\footnote{Refer to \ref{sec:appendix} for detailed definitions of Lipschitz continuous, traceable trajectory, and clearance.}, there exists some cutoff resolution for which \rcsnr will find a solution in finite time.
\end{thrm}
\vspace{2mm}

The third part of our proof, described in Sec.~\ref{subsec:node-rejection} of \ref{sec:appendix}, shows that even with similar node rejection, the basic version of our algorithm \rcsb still finds a solution when several conditions are satisfied.
Here we adapt the proofs provided by Cheng and LaValle~\cite[Thm. 5.2]{Cheng2002_ICRA}.
In their proof, a fixed control period is assumed for every motion primitive.
Thus, we need to incorporate the machinery developed in the second part of our proof (Sec.~\ref{subsec:fixed-length-primitives} of \ref{sec:appendix}) and obtain the following result (again, stated informally to avoid using notations defined only in \ref{sec:appendix}).

\begin{thrm}[Resolution completeness with similar-node rejection]
\label{thrm:resolution_complete2}
    Let $\Delta = (\X, \W_{\rm obs}, \mathbf{x}_{\rm start}, p_{\rm goal}, \tau, \ell_{\max}, \kappa_{\max})$ be a steerable needle motion planning problem.
    Under the assumption that the system is Lipschitz continuous and there exists a traceable solution that has sufficient clearance, there exists some cutoff resolution~$\{\delta\ell_{\min}, \delta\theta_{\min}\}$
    and some radius for similar node rejection~$d_{\rm sim}$ (which is a function of $\tau, \ell_{\max}$ and $\delta\ell_{\min})$
    for which \rcsb will find a solution in finite time.
\end{thrm}
\vspace{2mm}

In the final part of the proof (Sec.~\ref{subsec:implementation-details2} of \ref{sec:appendix}), we show that none of the implementation details we use to improve the algorithm's efficiency hinder the theoretical guarantees of \rcsb.
\section{Results}
\label{sec:results}

We evaluate our new resolution-complete motion planner for steerable needles using scenarios based on the medical task of lung biopsy.
Lung cancer is the deadliest form of cancer in the United States, killing over 150,000 Americans each year \cite{ACS2016}. Early diagnosis is critical for patient survival, and biopsy of suspicious nodules is required for diagnosis.
Steerable needles deployed from bronchoscopes have the potential to safely and accurately reach nodules throughout the lung for biopsy and localized treatment~\cite{Kuntz2016_Hamlyn,Swaney2017_JMRR}.
We illustrate in Fig.~\ref{fig:cover} a volumetric model of the relevant anatomy segmented from a CT scan \cite{Fu2018_IROS}.
In this procedure, the steerable needle is deployed from a bronchoscope inside the lung and must steer from the start pose just outside a bronchial tube (the furthest pose reachable by the bronchoscope) to the nodule while avoiding anatomical obstacles that include the large blood vessels, the bronchial tubes, and the lung boundary.

\begin{figure}
    \centering
    \includegraphics[width=0.95\linewidth]{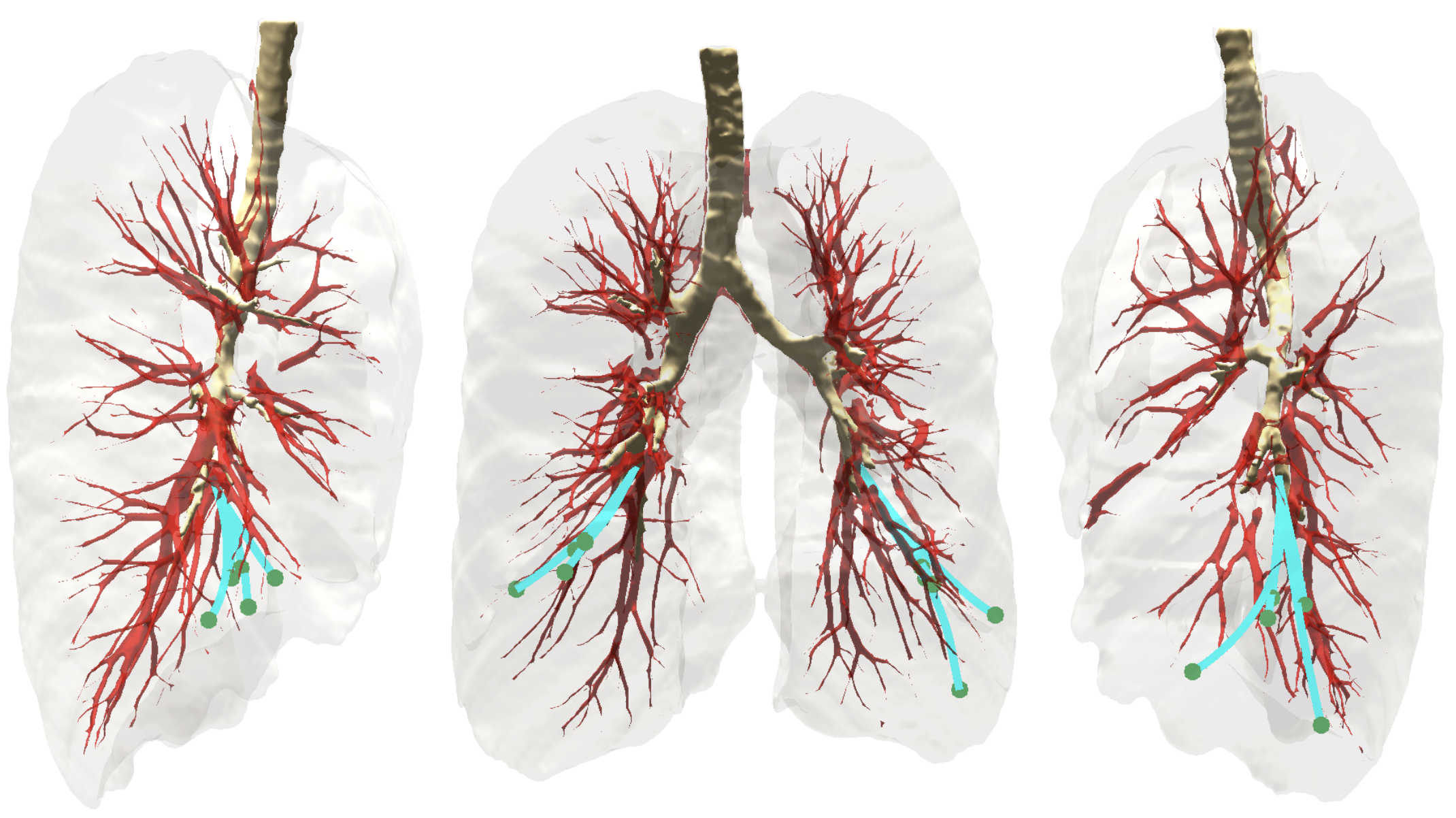}
    \caption{Three views of the the lung environment.
    The needle steers to targets (green) while avoiding anatomical obstacles including large blood vessels (red), bronchial tubes (brown), and the lung boundary (gray).
    We also show~10 of the 500 test cases in which the steerable needle must deploy from the bronchoscope's tip in the bronchial tube to the nodule in the lung parenchyma. For these example test cases, we show plans computed by \rcs (cyan). 
    }
    \label{fig:anatomy}
\vspace{-3mm}
\end{figure}

To create test cases, we randomly sampled $50$ collision-free start configurations along the bronchial tube walls (i.e., points reachable by the bronchoscope from which the steerable needle can be deployed), each with~$10$ reachable goal points in the lung parenchyma (i.e., points in the tissue of the lung outside the bronchial tubes in which nodules requiring biopsy may occur), totaling 500 test cases (see Fig.~\ref{fig:anatomy} for 10  plans computed by \rcs).
To avoid skewing the data with trivial test cases, we discarded test cases where the start configuration can be connected directly to the goal point with a collision-free arc. 
Additionally, we also disallowed test cases where there are obstacles directly in front of the start configuration deeming the problem unsolvable.
Finally, note that it is not guaranteed that a valid plan exists for a test case.

We consider a steerable needle with a maximum curvature of $\kappa_{\max} = 0.01 ({\rm mm}^{-1})$, 
device diameter of $2{\rm mm}$,
and maximum insertion length of $100 {\rm mm}$.
The simulated workspace was reconstructed from a preoperative chest CT scan where~$\W_{\rm obs}$ is a point cloud representing the anatomical obstacles described above.
We use a collision-checking resolution of $0.5{\rm mm}$ 
and a goal tolerance of~$\tau = 1.0{\rm mm}$.

We compared in simulation the variants of \rcs with two steerable needle planners: an \rrt-based planner~\cite{Kuntz2015_IROS,Patil2014_TRO} and \aft~\cite{Liu2016_RAL, Pinzi2019_IJCARS}.
While the original \aft algorithm is GPU accelerated, here we present results for our CPU-based implementation and only focus on the feasibility of the method and not on the computing times (we let \aft run until it terminates).
Similar to~\cite{Pinzi2019_IJCARS}, we define the cost function for \aft as
\begin{equation}
\label{eq:cost}
    {\rm Cost(\sigma)} = \ell(\sigma)/\ell_{\max} + \|\sigma(1) - p_{\rm goal}\|_2/\tau,
\end{equation}
which accounts both for insertion length and final tip error.
We also ran a search-based planner denoted as \algname{SINGLE\_RES} that includes all optimizations of RCS mentioned in Sec.~\ref{subsec:details} but that uses only the finest resolution (with no multiple resolutions).
For additional details about the parameters used for each planner, see~\ref{sec:appendix-planner-params}.
All experiments were run on a dual 2.1GHz 16-core Intel Xeon Silver 4216 CPU and 100GB of RAM.
Code for our proposed method is available on GitHub~\cite{Fu2021_GitHub}.

\begin{figure}
    \centering
    \includegraphics[width=\linewidth]{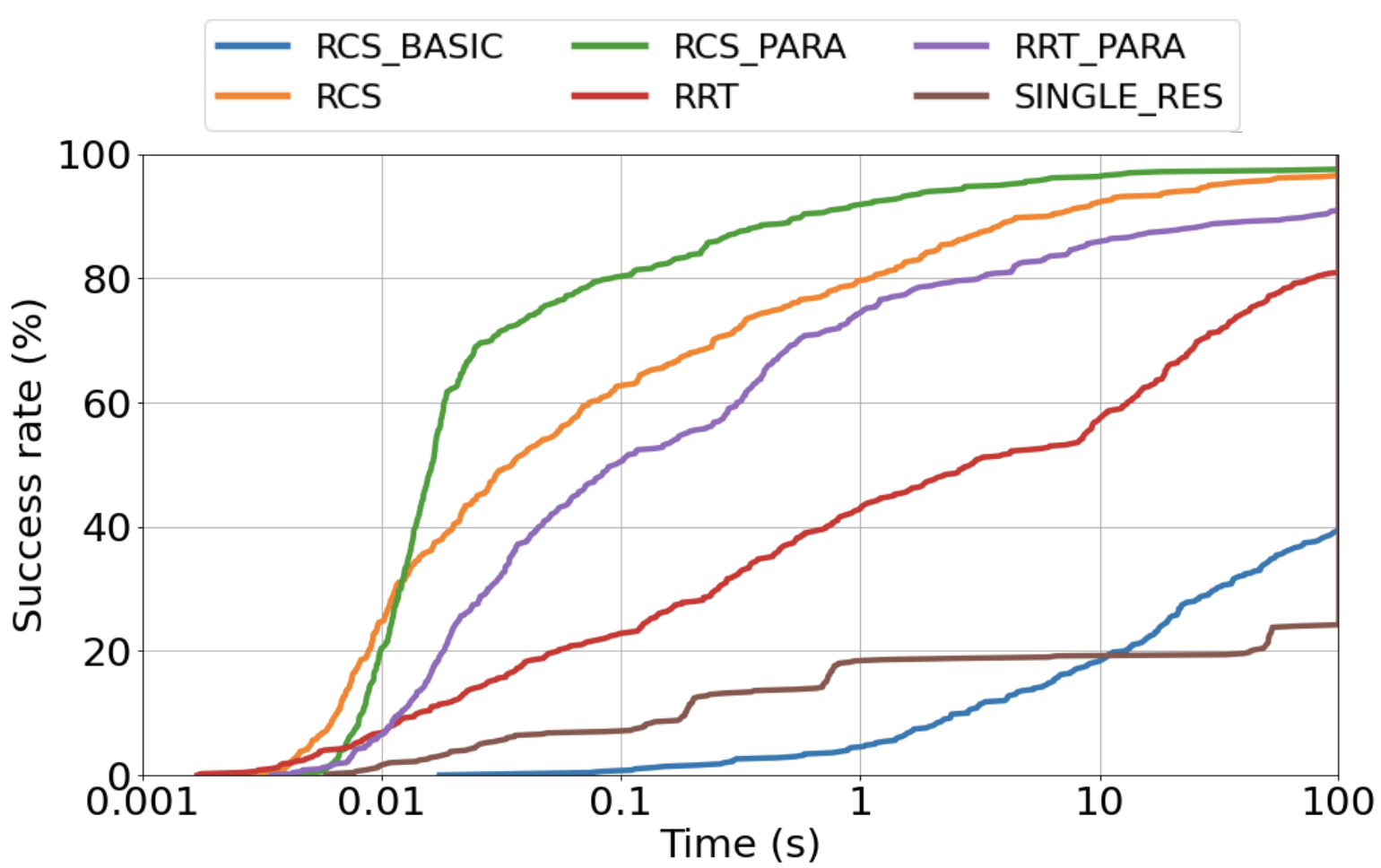}
    \caption{Success rate as a function of time for \rcs and \rrt.
    }
    \label{fig:success_rate}
\vspace{-5mm}
\end{figure}

We now present results pertaining to the success rate of the different algorithms.
In our setting, the success rate is the ratio of solved cases among all 500 test cases.
For \rcs, \rrt, and their variants, each planner was allowed $100$ seconds.
The results are shown in Fig.~\ref{fig:success_rate}.
First, among \rcs variants, \rcs performed much better than \rcsb, indicating the first three optimizations introduced in Sec.~\ref{subsec:details} dramatically improved the efficiency of the algorithm.
Furthermore, except for the obvious overhead effect in the early stage ($< 10 {\rm ms}$), \rcspara achieved even better performance.
The single-resolution planner \algname{SINGLE\_RES} only achieved a~$24.2\%$ success rate, suggesting that the multi-resolution approach in \rcs variants is valuable.
Second, the single-threaded \rcs achieved better performance than the single-threaded \rrt and multi-threaded \rrtpara.
From the perspective of running time, 
\rcspara's average running time for solved cases is $0.43$ seconds,
and it took $0.83$ seconds to reach a success rate of~$91.2\%$, which is roughly $120$ times faster than \rrtpara.

Since we only had a CPU-based version of the \aft algorithm, we do not compare success rate over time. Instead, we compare the success rate when \rcs runs for $100$ seconds and \aft finishes two tree refinements.
Additionally, as \aft produces many paths while optimizing a cost function that does not necessarily favor paths with minimal goal tolerance~(Eq.~\ref{eq:cost}), we chose the one with the minimal goal tolerance (not with the minimal cost) for success rate analysis.
The 5-level \aft achieved a success rate of~$65.8\%$, with many of the failures due to the computed paths not satisfying the maximum allowed targeting error of $\tau = 1 {\rm mm}$.

For additional experiments evaluating the quality of the plans produced by each planner, see~\ref{sec:appendix-experiments}.
\section{Conclusion \& future work}
\label{sec:conclusion}

In this paper, we took  an  important  step  toward  creating  a  certifiable  motion planner  for  steerable  needles. Specifically, we introduced a resolution-complete planner that dramatically outperforms state-of-the-art needle planners in a clinically inspired simulation. This was achieved by carefully designing motion primitives and applying domain-specific optimizations. We formally showed that the planner is resolution complete, which means that under some mild assumptions on the system  and  the  solution, the planner, in finite time, is guaranteed to find a  plan  as  long  as  the  problem  admits  a  qualified  solution. 

We view this work as an \emph{algorithmic foundation} required to obtain certifiable motion planning for steerable needles. Our planner is the first resolution-complete planner for steerable needles, but more work remains. Our analysis showed that, under some mild assumptions, when a qualified solution exists, \emph{if} the cutoff resolution is fine enough and the path has \emph{some} clearance (distance from the obstacles), the algorithm will find it.
However, it would be valuable for medical applications to provide the precise relation between the system's controls and this cutoff resolution. Subsequently, we need to provide the precise relation between this cutoff resolution (i.e., what does it mean to be ``fine enough'') and the clearance of paths (i.e., what does it mean ``some clearance''?). Future work will use this foundation to compute the relation between the aforementioned parameters in order to give physicians certifiable software for motion planning for steerable needles.

We believe that the algorithmic foundations laid out in this work will also allow us to provide guarantees on the \emph{quality of the solution}---a critical requirement in our medical domain. Here, trajectory quality can correspond to minimizing damage to tissue, the time the patient is under anaesthesia, and more (see~\cite{Bentley2021_CORR} and references within).
Consequently, we plan to revisit the way nodes are ordered in our priority queue (recall that now they are ordered according to their rank) in order to provide optimality (or near-optimality) guarantees.

\section*{Acknowledgments}
This research was supported in part by the U.S.\ National Institutes of Health (NIH) under award R01EB024864, the Israeli Ministry of Science \& Technology (MOST) by grant No.\ 102583 and 2028142, and the United States-Israel Binational Science Foundation (BSF) by grant No.\ 1018193.

We thank
Janine Hoelscher,
Inbar Fried,
Maxwell Emerson, 
Tayfun Efe Ertop, 
Margaret Rox, 
Josephine Granna, 
Alan Kuntz, 
Jason A.\ Akulian, 
and Robert J.\ Webster III
for their discussions on steerable needles for lung applications.

\bibliographystyle{plainnat}
\bibliography{IEEEabrv,refabrv,references}

\begin{thebibliography}{54}
\providecommand{\natexlab}[1]{#1}
\providecommand{\url}[1]{\texttt{#1}}
\expandafter\ifx\csname urlstyle\endcsname\relax
  \providecommand{\doi}[1]{doi: #1}\else
  \providecommand{\doi}{doi: \begingroup \urlstyle{rm}\Url}\fi

\bibitem[NDI()]{NDI}
Aurora - {NDI}.
\newblock \url{https://www.ndigital.com/products/aurora/}.
\newblock Accessed: 2021-02-28.

\bibitem[Abolhassani et~al.(2007)Abolhassani, Patel, and
  Moallem]{Abolhassani2007_MEP}
Niki Abolhassani, Rajni Patel, and Mehrdad Moallem.
\newblock Needle insertion into soft tissue: A survey.
\newblock \emph{Medical Engineering \& Physics}, 29\penalty0 (4):\penalty0
  413--431, 2007.

\bibitem[Alterovitz et~al.(2005)Alterovitz, Goldberg, and
  Okamura]{Alterovitz2005_ICRA}
Ron Alterovitz, Ken Goldberg, and Allison Okamura.
\newblock Planning for steerable bevel-tip needle insertion through {2D} soft
  tissue with obstacles.
\newblock In \emph{{{IEEE} Int. Conf. Robotics and Automation ({ICRA})}}, pages
  1640--1645. IEEE, 2005.

\bibitem[Alterovitz et~al.(2007)Alterovitz, Sim{\'e}on, and
  Goldberg]{Alterovitz2007_RSS}
Ron Alterovitz, Thierry Sim{\'e}on, and Ken Goldberg.
\newblock The stochastic motion roadmap: A sampling framework for planning with
  markov motion uncertainty.
\newblock In \emph{{Robotics: Science and Systems (RSS)}}, 2007.

\bibitem[{American Cancer Society}(2016)]{ACS2016}
{American Cancer Society}.
\newblock {Cancer Facts \& Figures}.
\newblock Technical report, American Cancer Society, 2016.

\bibitem[Asadian et~al.(2011)Asadian, Kermani, and Patel]{Asadian2011_JINT}
Ali Asadian, Mehrdad~R Kermani, and Rajni~V Patel.
\newblock Robot-assisted needle steering using a control theoretic approach.
\newblock \emph{{J. Intelligent and Robotic Systems}}, 62\penalty0
  (3):\penalty0 397--418, 2011.

\bibitem[Barraquand and Latombe(1991)]{Barraquand1991_IJRR}
Jerome Barraquand and Jean-Claude Latombe.
\newblock Robot motion planning: A distributed representation approach.
\newblock \emph{{Int. J. Robotics Research (IJRR)}}, 10\penalty0 (6):\penalty0
  628--649, 1991.

\bibitem[Barraquand and Latombe(1993)]{Barraquand1993_Algorithmica}
J{\'e}r{\^o}me Barraquand and Jean-Claude Latombe.
\newblock Nonholonomic multibody mobile robots: Controllability and motion
  planning in the presence of obstacles.
\newblock \emph{Algorithmica}, 10\penalty0 (2):\penalty0 121--155, 1993.

\bibitem[Bentley et~al.(2021)Bentley, Rucker, Reddy, Salzman, and
  Kuntz]{Bentley2021_CORR}
Michael Bentley, Caleb Rucker, Chakravarthy Reddy, Oren Salzman, and Alan
  Kuntz.
\newblock A novel shaft-to-tissue force model for safer motion planning of
  steerable needles.
\newblock \emph{{Computing Research Repository (CoRR)}}, abs/2101.02246, 2021.

\bibitem[Bernardes et~al.(2012)Bernardes, Adorno, Poignet, and
  Borges]{Bernardes2012_ICRA}
Mariana~C Bernardes, Bruno~V Adorno, Philippe Poignet, and Geovany~A Borges.
\newblock Semi-automatic needle steering system with robotic manipulator.
\newblock In \emph{{{IEEE} Int. Conf. Robotics and Automation ({ICRA})}}, pages
  1595--1600. IEEE, 2012.

\bibitem[Cheng and LaValle(2002)]{Cheng2002_ICRA}
Peng Cheng and Steven~M LaValle.
\newblock Resolution complete rapidly-exploring random trees.
\newblock In \emph{{{IEEE} Int. Conf. Robotics and Automation ({ICRA})}},
  volume~1, pages 267--272. IEEE, 2002.

\bibitem[Cowan et~al.(2011)Cowan, Goldberg, Chirikjian, Fichtinger, Alterovitz,
  Reed, Kallem, Park, Misra, and Okamura]{Cowan2011_Chapter}
Noah~J Cowan, Ken Goldberg, Gregory~S Chirikjian, Gabor Fichtinger, Ron
  Alterovitz, Kyle~B Reed, Vinutha Kallem, Wooram Park, Sarthak Misra, and
  Allison~M Okamura.
\newblock {Robotic needle steering: design, modeling, planning, and image
  guidance}.
\newblock In Jacob Rosen, Blake Hannaford, and Richard~M Satava, editors,
  \emph{Surgical Robotics: System Applications and Visions}, chapter~23, pages
  557--582. Springer, 2011.

\bibitem[DiMaio and Salcudean(2003)]{Dimaio2003_TRA}
Simon~P DiMaio and Septimiu~E Salcudean.
\newblock Needle insertion modeling and simulation.
\newblock \emph{{{IEEE} Trans. Robotics and Automation}}, 19\penalty0
  (5):\penalty0 864--875, 2003.

\bibitem[Du et~al.(2019)Du, Kim, Salzman, and Likhachev]{Du2019_IROS}
Wei Du, Sung-Kyun Kim, Oren Salzman, and Maxim Likhachev.
\newblock Escaping local minima in search-based planning using soft duplicate
  detection.
\newblock In \emph{{IEEE/RSJ Int. Conf. Intelligent Robots and Systems
  (IROS)}}, pages 2365--2371. IEEE, 2019.

\bibitem[Duindam et~al.(2010)Duindam, Xu, Alterovitz, Sastry, and
  Goldberg]{Duindam2010_IJRR}
Vincent Duindam, Jijie Xu, Ron Alterovitz, Shankar Sastry, and Ken Goldberg.
\newblock Three-dimensional motion planning algorithms for steerable needles
  using inverse kinematics.
\newblock \emph{{Int. J. Robotics Research (IJRR)}}, 29\penalty0 (7):\penalty0
  789--800, 2010.

\bibitem[Favaro et~al.(2018)Favaro, Cerri, Galvan, Baena, and
  De~Momi]{Favaro2018_ICRA}
Alberto Favaro, Leonardo Cerri, Stefano Galvan, Ferdinando Rodriguez~Y Baena,
  and Elena De~Momi.
\newblock Automatic optimized {3D} path planner for steerable catheters with
  heuristic search and uncertainty tolerance.
\newblock In \emph{{{IEEE} Int. Conf. Robotics and Automation ({ICRA})}}, pages
  9--16. IEEE, 2018.

\bibitem[Frazzoli et~al.(2002)Frazzoli, Dahleh, and Feron]{Frazzoli2002_JGCD}
Emilio Frazzoli, Munther~A Dahleh, and Eric Feron.
\newblock Real-time motion planning for agile autonomous vehicles.
\newblock \emph{Journal of Guidance, Control, and Dynamics}, 25\penalty0
  (1):\penalty0 116--129, 2002.

\bibitem[Fu et~al.(2018)Fu, Kuntz, Webster~III, and Alterovitz]{Fu2018_IROS}
Mengyu Fu, Alan Kuntz, Robert~J Webster~III, and Ron Alterovitz.
\newblock Safe motion planning for steerable needles using cost maps
  automatically extracted from pulmonary images.
\newblock In \emph{{IEEE/RSJ Int. Conf. Intelligent Robots and Systems
  (IROS)}}, pages 4942--4949. IEEE, 2018.

\bibitem[Fu et~al.(2021)Fu, Salzman, and Alterovitz]{Fu2021_GitHub}
Mengyu Fu, Oren Salzman, and Ron Alterovitz.
\newblock {steerable-needle-planner}.
\newblock \url{https://github.com/UNC-Robotics/steerable-needle-planner}, 2021.
\newblock Accessed: 2021-6-9.

\bibitem[Hauser et~al.(2009)Hauser, Alterovitz, Chentanez, Okamura, and
  Goldberg]{Hauser2009_RSS}
K.~Hauser, R.~Alterovitz, N.~Chentanez, A.~Okamura, and K.~Goldberg.
\newblock Feedback control for steering needles through 3{D} deformable tissue
  using helical paths.
\newblock In \emph{Proceedings of Robotics: Science and Systems}, Seattle, USA,
  June 2009.

\bibitem[Hauser(2015)]{Hauser2015_ICRA}
Kris Hauser.
\newblock Lazy collision checking in asymptotically-optimal motion planning.
\newblock In \emph{{{IEEE} Int. Conf. Robotics and Automation ({ICRA})}}, pages
  2951--2957, 2015.

\bibitem[Ichnowski and Alterovitz(2019)]{Ichnowski2019_ICRA}
Jeffrey Ichnowski and Ron Alterovitz.
\newblock Motion planning templates: A motion planning framework for robots
  with low-power {CPU}s.
\newblock In \emph{{{IEEE} Int. Conf. Robotics and Automation ({ICRA})}}, pages
  612--618. IEEE, 2019.

\bibitem[Islam et~al.(2019)Islam, Salzman, and Likhachev]{Islam2019_ICAPS}
Fahad Islam, Oren Salzman, and Maxim Likhachev.
\newblock Provable indefinite-horizon real-time planning for repetitive tasks.
\newblock In \emph{{Int. Conf. Automated Planning and Scheduling (ICAPS)}},
  volume~29, pages 716--724, 2019.

\bibitem[Islam et~al.(2020)Islam, Vemula, Kim, Dornbush, Salzman, and
  Likhachev]{Islam2020_ICRA}
Fahad Islam, Anirudh Vemula, Sung-Kyun Kim, Andrew Dornbush, Oren Salzman, and
  Maxim Likhachev.
\newblock Planning, learning and reasoning framework for robot truck unloading.
\newblock In \emph{{{IEEE} Int. Conf. Robotics and Automation ({ICRA})}}, pages
  5011--5017. IEEE, 2020.

\bibitem[Karaman and Frazzoli(2011)]{Karaman2011_IJRR}
Sertac Karaman and Emilio Frazzoli.
\newblock Sampling-based algorithms for optimal motion planning.
\newblock \emph{{Int. J. Robotics Research (IJRR)}}, 30\penalty0 (7):\penalty0
  846--894, 2011.

\bibitem[Kirkpatrick et~al.(2011)Kirkpatrick, Kostitsyna, and
  Polishchuk]{Kirkpatrick2011_CCCG}
David~G. Kirkpatrick, Irina Kostitsyna, and Valentin Polishchuk.
\newblock Hardness results for two-dimensional curvature-constrained motion
  planning.
\newblock In \emph{Canadian Conference on Computational Geometry (CCCG)}, 2011.

\bibitem[Kleinbort et~al.(2018)Kleinbort, Solovey, Littlefield, Bekris, and
  Halperin]{Kleinbort2018_RAL}
Michal Kleinbort, Kiril Solovey, Zakary Littlefield, Kostas~E Bekris, and Dan
  Halperin.
\newblock Probabilistic completeness of {RRT} for geometric and kinodynamic
  planning with forward propagation.
\newblock \emph{{IEEE Robotics and Automation Letters}}, 4\penalty0
  (2):\penalty0 x--xvi, 2018.

\bibitem[Ko et~al.(2011)Ko, Frasson, and y~Baena]{Ko2011_TRO}
Seong~Young Ko, Luca Frasson, and Ferdinando~Rodriguez y~Baena.
\newblock Closed-loop planar motion control of a steerable probe with a
  “programmable bevel” inspired by nature.
\newblock \emph{{{IEEE} Trans. Robotics}}, 27\penalty0 (5):\penalty0 970--983,
  2011.

\bibitem[Kuntz et~al.(2016)Kuntz, Swaney, Mahoney, Feins, Lee, Webster~III, and
  Alterovitz]{Kuntz2016_Hamlyn}
A~Kuntz, P~J Swaney, A~Mahoney, R~H Feins, Y~Z Lee, Robert~J Webster~III, and
  Ron Alterovitz.
\newblock {Toward transoral peripheral lung access: Steering
  bronchoscope-deployed needles through porcine lung tissue}.
\newblock In \emph{Hamlyn Symposium on Medical Robotics}, pages 9--10, 2016.

\bibitem[Kuntz et~al.(2015)Kuntz, Torres, Feins, Webster~III, and
  Alterovitz]{Kuntz2015_IROS}
Alan Kuntz, Luis~G Torres, Richard~H Feins, Robert~J Webster~III, and Ron
  Alterovitz.
\newblock Motion planning for a three-stage multilumen transoral lung access
  system.
\newblock In \emph{{IEEE/RSJ Int. Conf. Intelligent Robots and Systems
  (IROS)}}, pages 3255--3261. IEEE, 2015.

\bibitem[LaValle(1998)]{Lavalle1998}
Steven~M LaValle.
\newblock Rapidly-exploring random trees: A new tool for path planning.
\newblock 1998.

\bibitem[LaValle(2006)]{Lavalle2006_BOOK}
Steven~M LaValle.
\newblock \emph{Planning algorithms}.
\newblock Cambridge university press, 2006.

\bibitem[Lindemann and LaValle(2006)]{Lindemann2006_ICRA}
Stephen~R Lindemann and Steven~M LaValle.
\newblock Multiresolution approach for motion planning under differential
  constraints.
\newblock In \emph{{{IEEE} Int. Conf. Robotics and Automation ({ICRA})}}, pages
  139--144. IEEE, 2006.

\bibitem[Liu et~al.(2016)Liu, Garriga-Casanovas, Secoli, and
  y~Baena]{Liu2016_RAL}
Fangde Liu, Arnau Garriga-Casanovas, Riccardo Secoli, and Ferdinando~Rodriguez
  y~Baena.
\newblock Fast and adaptive fractal tree-based path planning for programmable
  bevel tip steerable needles.
\newblock \emph{{IEEE Robotics and Automation Letters}}, 1\penalty0
  (2):\penalty0 601--608, 2016.

\bibitem[Ljungqvist et~al.(2017)Ljungqvist, Evestedt, Cirillo, Axehill, and
  Holmer]{Ljungqvist2017_IVS}
Oskar Ljungqvist, Niclas Evestedt, Marcello Cirillo, Daniel Axehill, and Olov
  Holmer.
\newblock Lattice-based motion planning for a general 2-trailer system.
\newblock In \emph{{{IEEE} Intelligent Vehicles Symposium (IV)}}, pages
  819--824. IEEE, 2017.

\bibitem[Mandalika et~al.(2019)Mandalika, Choudhury, Salzman, and
  Srinivasa]{Mandalika2019_ICAPS}
Aditya Mandalika, Sanjiban Choudhury, Oren Salzman, and Siddhartha~S.
  Srinivasa.
\newblock Generalized lazy search for robot motion planning: Interleaving
  search and edge evaluation via event-based toggles.
\newblock In \emph{{Int. Conf. Automated Planning and Scheduling (ICAPS)}},
  pages 745--753, 2019.

\bibitem[Minhas et~al.(2007)Minhas, Engh, Fenske, and Riviere]{Minhas2007_EMBC}
Davneet~S Minhas, Johnathan~A Engh, Michele~M Fenske, and Cameron~N Riviere.
\newblock Modeling of needle steering via duty-cycled spinning.
\newblock In \emph{{Annual International Conference of the {IEEE} Engineering
  in Medicine and Biology Society (EMBC)}}, pages 2756--2759. IEEE, 2007.

\bibitem[Okazawa et~al.(2005)Okazawa, Ebrahimi, Chuang, Salcudean, and
  Rohling]{Okazawa2005_ITM}
Stephen Okazawa, Richelle Ebrahimi, Jason Chuang, Septimiu~E Salcudean, and
  Robert Rohling.
\newblock Hand-held steerable needle device.
\newblock \emph{{{IEEE/ASME} Trans. Mechatronics}}, 10\penalty0 (3):\penalty0
  285--296, 2005.

\bibitem[Park et~al.(2005)Park, Kim, Zhou, Cowan, Okamura, and
  Chirikjian]{Park2005_ICRA}
Wooram Park, Jin~Seob Kim, Yu~Zhou, Noah~J Cowan, Allison~M Okamura, and
  Gregory~S Chirikjian.
\newblock {Diffusion-based motion planning for a nonholonomic flexible needle
  model}.
\newblock In \emph{Proc. IEEE Int. Conf. Robotics and Automation (ICRA)}, pages
  4611--4616, April 2005.

\bibitem[Patil et~al.(2014)Patil, Burgner, Webster~III, and
  Alterovitz]{Patil2014_TRO}
Sachin Patil, Jessica Burgner, Robert~J Webster~III, and Ron Alterovitz.
\newblock Needle steering in 3{D} via rapid replanning.
\newblock \emph{{{IEEE} Trans. Robotics}}, 30\penalty0 (4):\penalty0 853--864,
  2014.

\bibitem[Pinzi et~al.(2019)Pinzi, Galvan, and y~Baena]{Pinzi2019_IJCARS}
Marlene Pinzi, Stefano Galvan, and Ferdinando~Rodriguez y~Baena.
\newblock The adaptive hermite fractal tree ({AHFT}): a novel surgical 3{D}
  path planning approach with curvature and heading constraints.
\newblock \emph{{Int. J. Computer Assisted Radiology and Surgery}}, 14\penalty0
  (4):\penalty0 659--670, 2019.

\bibitem[Pivtoraiko and Kelly(2011)]{Pivtoraiko2011_IROS}
Mihail Pivtoraiko and Alonzo Kelly.
\newblock Kinodynamic motion planning with state lattice motion primitives.
\newblock In \emph{{IEEE/RSJ Int. Conf. Intelligent Robots and Systems
  (IROS)}}, pages 2172--2179. IEEE, 2011.

\bibitem[Qi et~al.(2014)Qi, Liu, Seneviratne, and Althoefer]{Qi2014_EMBC}
Peng Qi, Hongbin Liu, Lakmal Seneviratne, and Kaspar Althoefer.
\newblock Towards kinematic modeling of a multi-{DOF} tendon driven robotic
  catheter.
\newblock In \emph{{Annual International Conference of the {IEEE} Engineering
  in Medicine and Biology Society (EMBC)}}, pages 3009--3012. IEEE, 2014.

\bibitem[Reed et~al.(2011)Reed, Majewicz, Kallem, Alterovitz, Goldberg, Cowan,
  and Okamura]{Reed2011_RAM}
Kyle~B Reed, Ann Majewicz, Vinutha Kallem, Ron Alterovitz, Ken Goldberg, Noah~J
  Cowan, and Allison~M Okamura.
\newblock Robot-assisted needle steering.
\newblock \emph{{{IEEE} Robotics and Automation Magazine}}, 18\penalty0
  (4):\penalty0 35--46, 2011.

\bibitem[Rucker et~al.(2013)Rucker, Das, Gilbert, Swaney, Miga, Sarkar, and
  Webster~III]{Rucker2013_TRO}
D~Caleb Rucker, Jadav Das, Hunter~B Gilbert, Philip~J Swaney, Michael~I Miga,
  Nilanjan Sarkar, and Robert~J Webster~III.
\newblock {Sliding mode control of steerable needles}.
\newblock \emph{IEEE Trans. Robotics}, 29\penalty0 (5):\penalty0 1289--1299,
  2013.

\bibitem[Secoli and y~Baena(2016)]{Secoli2016_BIOROB}
Riccardo Secoli and Ferdinando~Rodriguez y~Baena.
\newblock Adaptive path-following control for bio-inspired steerable needles.
\newblock In \emph{{{IEEE} International Conference on Biomedical Robotics and
  Biomechatronics (BioRob)}}, pages 87--93. IEEE, 2016.

\bibitem[Seiler et~al.(2012)Seiler, Singh, Sukkarieh, and
  Durrant-Whyte]{Seiler2012_IJRR}
Konstantin~M Seiler, Surya~PN Singh, Salah Sukkarieh, and Hugh Durrant-Whyte.
\newblock Using {Lie} group symmetries for fast corrective motion planning.
\newblock \emph{{Int. J. Robotics Research (IJRR)}}, 31\penalty0 (2):\penalty0
  151--166, 2012.

\bibitem[Solovey(2020)]{Solovey2020_ARXIV}
Kiril Solovey.
\newblock Complexity of planning.
\newblock \emph{arXiv preprint arXiv:2003.03632v2 [cs.RO]}, 2020.

\bibitem[Sun et~al.(2015)Sun, Patil, and Alterovitz]{Sun2015_TRO}
Wen Sun, Sachin Patil, and Ron Alterovitz.
\newblock High-frequency replanning under uncertainty using parallel
  sampling-based motion planning.
\newblock \emph{{{IEEE} Trans. Robotics}}, 31\penalty0 (1):\penalty0 104--116,
  2015.

\bibitem[Swaney et~al.(2017)Swaney, Mahoney, Hartley, Remirez, Lamers, Feins,
  Alterovitz, and Webster~III]{Swaney2017_JMRR}
Philip~J Swaney, Arthur~W Mahoney, Bryan~I Hartley, Andria~A Remirez, Erik
  Lamers, Richard~H Feins, Ron Alterovitz, and Robert~J Webster~III.
\newblock Toward transoral peripheral lung access: Combining continuum robots
  and steerable needles.
\newblock \emph{Journal of Medical Robotics Research}, 2\penalty0
  (01):\penalty0 1750001, 2017.

\bibitem[Van Den~Berg et~al.(2010)Van Den~Berg, Patil, Alterovitz, Abbeel, and
  Goldberg]{Van2010_WAFR}
Jur Van Den~Berg, Sachin Patil, Ron Alterovitz, Pieter Abbeel, and Ken
  Goldberg.
\newblock {LQG}-based planning, sensing, and control of steerable needles.
\newblock In \emph{{Workshop on the Algorithmic Foundations of Robotics
  (WAFR)}}, pages 373--389. Springer, 2010.

\bibitem[Webster~III et~al.(2006)Webster~III, Kim, Cowan, Chirikjian, and
  Okamura]{Webster2006_IJRR}
Robert~J Webster~III, Jin~Seob Kim, Noah~J Cowan, Gregory~S Chirikjian, and
  Allison~M Okamura.
\newblock Nonholonomic modeling of needle steering.
\newblock \emph{{Int. J. Robotics Research (IJRR)}}, 25\penalty0
  (5-6):\penalty0 509--525, 2006.

\bibitem[Xu et~al.(2008)Xu, Duindam, Alterovitz, and Goldberg]{Xu2008_ICASE}
Jijie Xu, Vincent Duindam, Ron Alterovitz, and Ken Goldberg.
\newblock Motion planning for steerable needles in {3D} environments with
  obstacles using rapidly-exploring random trees and backchaining.
\newblock In \emph{{{IEEE} Int. Conf. Automation Science and Engineering}},
  pages 41--46. IEEE, 2008.

\bibitem[Yershov and LaValle(2010)]{Yershov2010_WAFR}
Dmitry~S Yershov and Steven~M LaValle.
\newblock Sufficient conditions for the existence of resolution complete
  planning algorithms.
\newblock In \emph{{Workshop on the Algorithmic Foundations of Robotics
  (WAFR)}}, pages 303--320. Springer, 2010.

\end{thebibliography}

\clearpage

\setcounter{section}{0}
\renewcommand{\thesection}{Appendix \Alph{section}}

\setcounter{subsection}{0}
\renewcommand{\thesubsection}{\Alph{subsection}}

\section{Resolution Completeness}
\label{sec:appendix}

\subsection{Preliminaries}
\label{subsec:preliminaries}
Before we state the different parts of our proof, we introduce some definitions.
Recall that a steerable needle motion planning problem is a tuple
$\Delta = (\X, \W_{\rm obs}, \mathbf{x}_{\rm start}, p_{\rm goal}, \tau, \ell_{\max}, \kappa_{\max})$
and that $\rho(\cdot)$ is a distance metric defined on $\X$ (Eq.~\ref{eq:metric}).
Finally, recall that $\A$ is the action space, which is the set of all valid motion primitives.
Throughout the proof, for some sequence of motion primitives~$M$, we will use 
$\mathbf{x} \oplus M$
to denote the resultant trajectory obtained by sequentially applying elements in $M$ to~$\mathbf{x}$.

\begin{dft}[Strong clearance]
\label{dft:clearance}
    Let $\sigma: [0, 1] \rightarrow \X$ be some trajectory.
    We say that
    $\sigma$ has strong $\gamma$-clearance if 
    $$\forall s \in [0, 1], \min_{\mathbf{x} \in \X_{\rm obs}} \rho(\sigma(s), \mathbf{x}) > \gamma,$$ 
    where $\X_{\rm obs} = {\rm cl}(\X \setminus \X_{\rm free})$ and ${\rm cl(\cdot)}$ is the closure of a set.
\end{dft}
\vspace{2mm}

\begin{dft}[Trajectory approximation]
\label{dft:approximation}
    Let~$\sigma: [0, 1] \rightarrow \X$ be some trajectory.
    We say that another trajectory~$\sigma'$ is an \textit{$\varepsilon$-approximation} of $\sigma$ if the following conditions are satisfied:
\begin{enumerate}[label=(\roman*)]
    \item boundary condition: $\forall s \in \{0, 1\}, \rho(\sigma(s), \sigma'(s)) < \varepsilon$;
    \item one-way Hausdorff distance:
    \begin{equation*}
        \max_{t \in [0,1]}\{\min_{s \in [0,1]}\rho(\sigma(s), \sigma'(t))\} < \varepsilon.
    \end{equation*}
\end{enumerate}
\end{dft}
\vspace{2mm}

\begin{dft}[Decomposable trajectory]
\label{dft:decomposable}
    Let~$\sigma: [0, 1] \rightarrow \X$ be some trajectory.
    We say that~$\sigma$ is decomposable if it can be decomposed into a finite sequence of motion primitives.
    Namely, there exists  $M_{\sigma} = \{\M_1,\dots,\M_n\} \subset \A$ such that $\sigma = \sigma(0) \oplus M_\sigma$.
\end{dft}
\vspace{2mm}

\begin{dft}[Traceable trajectory]
\label{dft:tracable}
    Let~$\sigma: [0, 1] \rightarrow \X$ be some trajectory.
    We say that~$\sigma$ is traceable if for any given $\varepsilon > 0$,
    there exists a decomposable trajectory that  is an $\varepsilon$-approximation of $\sigma$.
\end{dft}
\vspace{2mm}

Note that in the definitions of decomposable and traceable trajectories we allow any arbitrary set of motion primitives.
This allows us to decouple the planner's ability of finding a path using a given set of motion primitives with the expressiveness of the motion primitives.

\begin{dft}
\label{dft:action-distance}
    We define a distance metric on action space~$\A$ as
    the two-way Hausdorff distance between two resultant trajectories
    $\mathbf{x} \oplus \M_1$
    and
    $\mathbf{x} \oplus \M_2$.
    Formally, we have
    \begin{equation*}
    \begin{split}
        \rho_{\A}(\M_1, \M_2) = {\max}
        \Big\{
        & \max_{t \in [0,1]}\{
        \min_{s \in [0, 1]}
        \rho(\sigma_{\M_1}(s), \sigma_{\M_2}(t))\},\\
        & \max_{s \in [0,1]}\{
        \min_{t \in [0, 1]}
        \rho(\sigma_{\M_1}(s), \sigma_{\M_2}(t))
        \}
        \Big\},
    \end{split}
    \end{equation*}
    where 
    $\sigma_{\M_1} = \mathbf{x} \oplus \M_1$
    and
    $\sigma_{\M_2} = \mathbf{x} \oplus \M_2$.
    It is worth to note that changing $\mathbf{x}$ does not change the relative position between the two trajectories.
    Thus, without losing generality, we have 
    $\mathbf{x} = (p, q)$
    where
    $p = (0,0,0)$
    and
    $q = (1,0,0,0)$.
\end{dft}

\begin{dft}[Lipschitz continuous]
\label{dft:lipschitz}
    The system is Lipschitz continuous if
    $\forall \mathbf{x}_1, \mathbf{x}_2 \in \X, \forall \M_1, \M_2 \in \A$,
    \begin{equation*}
        \rho(\mathbf{x}_1 \oplus \M_1, \mathbf{x}_2 \oplus \M_2) \leq L_s(\rho(\mathbf{x}_1, \mathbf{x}_2) + \rho_{\A}(\M_1, \M_2)),
    \end{equation*}    
    where $L_s > 0$ is a constant.
\end{dft}
\vspace{2mm}

Finally, as we will see, it will be convenient to introduce the notion of a \textit{finest set} of motion primitives.
\begin{dft}[Finest set of motion primitives]
\label{dft:finest_set}
    Given a resolution $R = \{r_\ell, r_\theta\}$,
    and a set of curvatures~$K$,
    we define the \emph{finest set of motion primitives} as
    \begin{equation*}
        M_{\rm fs}(R,K) = 
        \bigg\{
         (\kappa, r_\ell, n\cdot r_\theta) 
         ~\Big|~
         \kappa \in K,
         n \in \left[ 
                    0, \left \lfloor\frac{2\pi}{r_{\theta}}\right \rfloor   
               \right] \subset \mathbb{Z}
        \bigg\}.
    \end{equation*}
\end{dft}

\subsection{Approximating curves with arbitrary curvatures}
\label{subsec:duty-cycling}
When a bevel-tip needle is inserted only, it follows a trajectory with curvature $\kappa_{\max}$.
When the needle is inserted while applying axial rotational velocity that is relatively larger than the insertion velocity, it follows a straight line (i.e., of curvature zero).
Minhans et al.~\cite{Minhas2007_EMBC} introduced the notion of  duty-cycling to approximate any curvature for bevel-tip steerable needles.
Roughly speaking, combining periods of needle spinning (i.e., zero-curvature trajectories) with periods of non-spinning (i.e., maximal-curvature trajectories) enables the needle to achieve any curvature up to the maximum needle curvature.
This idea is formalized in the following lemma.
\begin{lem}[Arbitrary curvature approximation using duty-cycling]
\label{lem:duty-cycling}
    Let~$\sigma$ be a decomposable trajectory
    and let~ $\varepsilon_d > 0$ be some real value.
    There exists a finite sequence of motion primitives $M_D$ in which every element has curvature  $\kappa \in \{0, \kappa_{\max}\}$
    such that the trajectory
    $\sigma(0) \oplus M_D$ is an $\varepsilon_d$-approximation of~$\sigma$.
\end{lem}

\vspace{2mm}
\emph{Proof sketch.}
Here, to explicitly show how the approximation factor is used. And to provide a more general discussion, we provide a proof from a geometric perspective (and not control-based as in the original work by Minhas et al.~\cite{Minhas2007_EMBC}).

The trajectory $\sigma$ is decomposable, thus there exists a sequence of motion primitives $M_{\sigma} = \{\M_1,\dots,\M_n\}$ such that 
$\sigma = \sigma(0) \oplus M_{\sigma}$ and each motion primitive $\M_i$ has arbitrary curvature $\kappa_i \in [0, \kappa_{\max}]$.
To approximate $\M_i$, we construct a sequence of motion primitives
$M_i = \{\M_i^{(1)},\dots,\M_i^{(n_i)}\}$ that satisfies
\begin{equation*}
\begin{split}
    &\M_i^{(1)}.\delta\theta = \M_i.\delta\theta, \\
    &\forall j \in [2, n_i], \M_i^{(j)}.\delta\theta = 0, \\
    &\forall j \in [1, n_i], \M_i^{(j)}.\kappa \in \{0, \kappa_{\max}\}.
\end{split}
\end{equation*}
Namely, the first motion primitive $\M_i^{(1)}$ ensures that both trajectories use the same curving plane (see Fig.~\ref{fig:primitive}) and the the rest of the sequence stays within this curving plane and approximates the (arbitrary) curvature $\kappa_i$.

\begin{figure}
    \centering
    \includegraphics[width=0.8\linewidth]{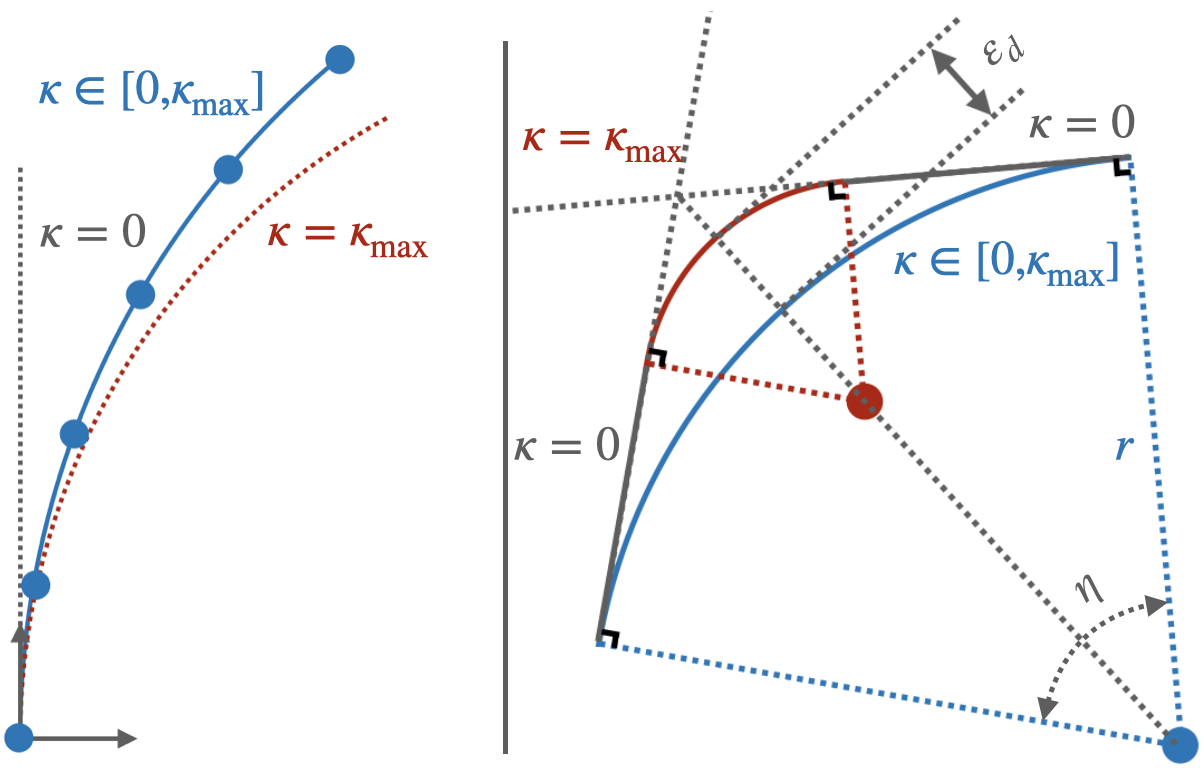}
    \caption{
    \textbf{Illustration of approximation with duty-cycling.}
    \textbf{Left:} Decompose~$\M_i$ into multiple segments with length~$\ell_i$.
    \textbf{Right:} Use three segments to approximate one segment of $\M_i$, where the segments have a curvature of $0, \kappa_{\max}, 0$, respectively.
    The one-way Hausdorff distance (marked as $\varepsilon_d$ in the figure) depends on $\ell_i$.
    For a given $\kappa_{\max}$, to approximate $\M_i$ (with curvature $\kappa$), the shorter $\ell_i$ is, the smaller $\varepsilon_d$ is.
    This is because
    $\varepsilon_d < r\cdot(1/{\rm cos}(0.5\eta) - 1)$,
    where $r = 1/\kappa$ is the radius of curvature and $\eta = \ell_i/r$ is the central angle.
    }
    \label{fig:duty-cycling}
\end{figure}

We then decompose $\M_i$ into small equal-length segments of length $\ell_i$ (except possibly the last segment) where the specific value of $\ell_i$ is chosen according to the value of  $\varepsilon_d $.
We then use three motion primitives to approximate each of these segments as illustrated in Fig.~\ref{fig:duty-cycling}.
It is not hard to see that 
(i)~the start and end configurations of $\M_i$ and $M_i$ are identical, and
(ii)~the one-way Hausdorff distance between $\M_i$ and each $\M_i^{(j)}$ is less than $\varepsilon_d $ if $\ell_i$ is carefully chosen.

Let  
$M_\sigma^{\varepsilon_d} = M_1 \cdot M_2 \cdot \ldots \cdot M_n$ be this sequence of all the newly constructed motion primitives.
Then it is straightforward that
$\sigma(0) \oplus M_\sigma^{\varepsilon_d}$ is an $\varepsilon_d$-approximation of $\sigma$.
$\hfill\blacksquare$

\subsection{Approximating curves using fixed-length primitives}
\label{subsec:fixed-length-primitives}

\begin{lem}[Fixed-resolution trajectory approximation]
\label{lem:resolution-existance}
    Let~$\sigma$ be a decomposable trajectory
    and let~ $\varepsilon_r > 0$ be some real value.
    If the system is Lipschitz continuous (Def.~\ref{dft:lipschitz}), 
    there exists a fine resolution $R(\sigma, \varepsilon_r) = \{r_{\ell}, r_{\theta}\}$
    and a finite sequence of motion primitives 
    $M_{R(\sigma, \varepsilon_r)}$
    such that  
    $\sigma(0) \oplus M_{R(\sigma, \varepsilon_r)}$
    is an $\varepsilon_r$-approximation of $\sigma$.
    Moreover $M_{R(\sigma, \varepsilon_r)} \subseteq \M_{\rm fs}(R(\sigma, \varepsilon_r), K_\sigma)$,
    where $K_\sigma$ is the set of curvatures that appear along $\sigma$.
\end{lem}

\vspace{2mm}
\textit{Proof sketch (adapted from~\cite[Appendix A]{Barraquand1991_IJRR}).}
The trajectory~$\sigma$ is decomposable, thus there exists a finite sequence of  motion primitives $M_\sigma = \{\M_1,\dots, \M_n\}$ such that $\sigma = \sigma(0) \oplus M_\sigma$.
Set $K_\sigma = \bigcup_i \M_i.\kappa$ to be the set of all curvatures that appear in $M_\sigma$.

To approximate each motion primitive~$\M_i$ using primitives from the finest set of motion primitives $\M_{\rm fs}(R(\sigma, \varepsilon_r), K_\sigma)$ (Def.~\ref{dft:finest_set}), we construct a sequence motion primitive~$M_i = \{\M_i^{(1)},\dots\M_i^{(n_i)}\}$,
where 
\begin{equation*}
    \begin{split}
        &\M_i^{(1)}.\delta\theta = k_i\cdot r_{\theta},\\ &\forall j \in [2, n_i], \M_i^{(j)}.\delta\theta = 0, \\
        &\forall j \in [1, n_i], \M_i^{(j)}.\kappa = \M_i.\kappa, M_i^{(j)}.\delta\ell = r_{\ell}.
    \end{split}
\end{equation*}

Similar to the sequence constructed for Lemma~\ref{lem:duty-cycling}, the first motion primitive $\M_i^{(1)}$ accounts for the curving plane (though here it can only be approximated)
and the the rest of the sequence stays within this curving plane and accounts for the length of the circular arc the trajectory follows in this plane.
Applying the sequence~$M_i$ is equivalent to applying one motion primitive
$\tilde{\M}_i = (\M_i.\kappa, n_i\cdot r_\ell, k_i\cdot r_\theta)$.
Thus, by carefully choosing~$r_\ell$ and~$r_\theta$, distance between 
$\M_i$
and
$\tilde{\M}_i$
(see Def.~\ref{dft:action-distance})
can be arbitrarily small.

\begin{figure}
    \centering
    \includegraphics[width=\linewidth]{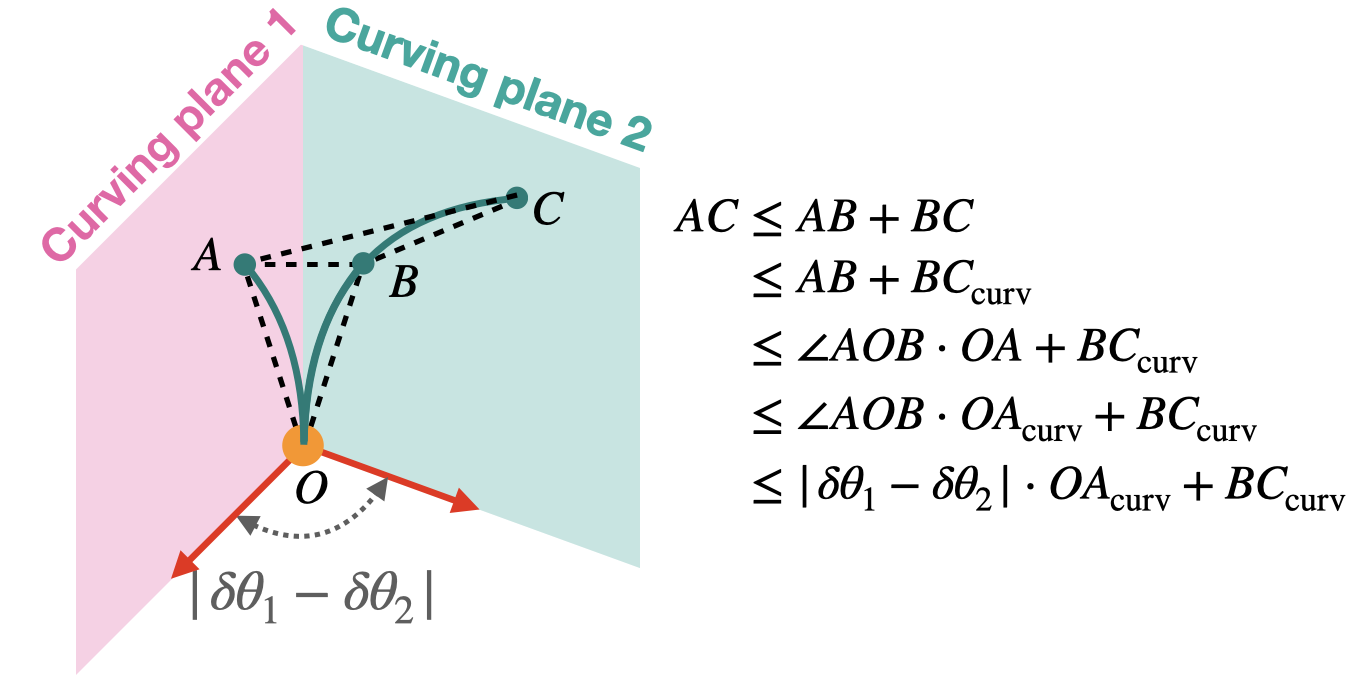}
    \caption{Illustration of the action distance between two motion primitives with the same curvature.
    Here the shorter motion primitive lies in curving plane 1, thus 
    ${\min}\{\delta\ell_1, \delta\ell_2\} = OA_{\rm curv}$
    and
    $|\delta\ell_1 - \delta\ell_2| = OC_{\rm curv} - OA_{\rm curv} = BC_{\rm curv}$.
    }
    \label{fig:action-distance}
\end{figure}

This is done for every motion primitive $\M_i$.
As $M$ is a finite sequence of size $n$, for any $\varepsilon > 0$ we can always find a fine-enough resolution $\{r_\ell, r_\theta\}$ that ensures that
$$\rho_{\A}(\M_i, \tilde{\M}_i) < \varepsilon, \forall i \in [1, n].$$
\noindent
This is because, given that both motion primitives have equal curvature, 
$\rho_{\A}(\M_1, \M_2) < |\delta\theta_1 - \delta\theta_2|\cdot{\min}\{\delta\ell_1, \delta\ell_2\} + |\delta\ell_1 - \delta\ell_2|$,
where
$\delta\ell_i = \M_i.\delta\ell$ and $\delta\theta_i = \M_i.\delta\theta$.
See Fig.~\ref{fig:action-distance} for illustration.

Since the system is Lipschitz continuous,
\begin{equation*}
\begin{split}
    &\rho(\sigma(0) \oplus \M_1 \dots \oplus \M_n, \sigma(0) \oplus \tilde{\M}_1 \dots \oplus \tilde{\M}_n) \\
    \leq &L_s(\rho(\sigma(0) \oplus \M_1 \dots \oplus \M_{n-1}, \sigma(0) \oplus \tilde{\M}_1 \dots \oplus \tilde{\M}_{n-1}) \\
    &+ \rho_{\A}(\M_n, \tilde{\M}_n) \\
    \leq & L_s^n\cdot \rho(\sigma(0), \sigma(0))
    + \sum_{i = 1}^{n}{L_s^{n-i+1} \cdot \rho_{\A}(\M_i, \tilde{\M}_i)}\\
    <& \varepsilon \cdot \frac{L_s(L_s^n - 1)}{L_s - 1}.
\end{split}
\end{equation*}
Thus, to ensure that $\sigma(0) \oplus \{\tilde{\M}_1, \dots, \tilde{\M}_n\}$ is an 
$\varepsilon_r$-approximation of $\sigma$, we only need to ensure that 
$\varepsilon \leq \frac{\varepsilon_r(L_s-1)}{L_s(L_s^n - 1)}$.
As both $n$ and $L_s$ are fixed, we can choose $\varepsilon$ to be as small as needed thus the desired fine resolution exists which concludes the proof.
$\hfill\blacksquare$

\begin{cor}
\label{cor:decomposition}
    Let~$\sigma$ be a traceable trajectory
    and let~ $\varepsilon > 0$ be some real value.
    If the system is Lipschitz continuous (Def.~\ref{dft:lipschitz}), 
    there exists a fine resolution $R(\sigma, \varepsilon) = \{r_{\ell}, r_{\theta}\}$
    and a finite sequence of motion primitives 
    $M_{R(\sigma, \varepsilon_r)}  \subseteq M_{\rm fs}(R(\sigma, \varepsilon), \{0, \kappa_{\max}\})$
    such that  
    $\sigma(0) \oplus M_{R(\sigma, \varepsilon_r)}$
    is an $\varepsilon$-approximation of $\sigma$.
\end{cor}

\vspace{2mm}
\textit{Proof sketch.}
Set $\varepsilon_t = \varepsilon_d = \varepsilon_r = \varepsilon / 3$.
According to Def.~\ref{dft:tracable},
there exists a decomposable trajectory $\sigma_t$ that is an $\varepsilon_t$-approximation of $\sigma$.
Moreover, according to Lemma~\ref{lem:duty-cycling}, 
there exists a finite sequence of motion primitives $M_D$ in which every element has curvature  $\kappa \in \{0, \kappa_{\max}\}$ such that the trajectory $\sigma_d = \sigma(0) \oplus M_D$ is an $\varepsilon_d$-approximation of~$\sigma_t$.

Note that by construction $\sigma_d$ is decomposable.
Thus, according to Lemma~\ref{lem:resolution-existance}, 
there exists a fine resolution $R(\sigma, \varepsilon_r) = \{r_{\ell}, r_{\theta}\}$ and a finite sequence of motion primitives  $M_{R(\sigma, \varepsilon_r)}$ such that $\sigma_r = \sigma(0) \oplus M_{R(\sigma, \varepsilon_r)}$ is an $\varepsilon_r$-approximation of $\sigma_d$.
Moreover, $M_{R(\sigma, \varepsilon_r)} \subseteq \M_{\rm fs}(R(\sigma, \varepsilon_r), \{0, \kappa_{\max}\})$ as the construction in the proof of Lemma~\ref{lem:resolution-existance} does not add new curvatures.

Finally, as $\varepsilon_d = \varepsilon_d = \varepsilon_r =  \varepsilon / 3$.
Then the trajectory~$\sigma_r$ is an $\varepsilon$-approximation of $\sigma$.
$\hfill\blacksquare$

\subsection{Resolution completeness (without similar node rejection)}

\begin{thm}[Resolution completeness of \rcsnr]
\label{thm:resolution_complete1}
    Let $\Delta = (\X, \W_{\rm obs}, \mathbf{x}_{\rm start}, p_{\rm goal}, \tau, \ell_{\max}, \kappa_{\max})$ be a steerable needle motion planning problem.
    If a solution to $\Delta$ is traceable, has strong $\gamma$-clearance for some $\gamma > 0$, and the system is Lipschitz continuous then there exists some cutoff resolution~$R_{\min}$ for which \rcsnr will find a solution in finite time.
\end{thm}

\vspace{2mm}
\textit{Proof sketch.}
Let $\sigma$ be a traceable solution with clearance~$\gamma$.
Following Cor.~\ref{cor:decomposition}, 
there exists 
 a fine resolution 
    $R(\sigma, \varepsilon) = \{r_{\ell}, r_{\theta}\}$
and 
 a finite sequence of motion primitives 
    $M_{R(\sigma, \varepsilon)}  \subseteq M_{\rm fs}(R(\sigma, \varepsilon_r), \{0, \kappa_{\max}\})$
such that  
 $\sigma(0) \oplus M_{R(\sigma, \varepsilon)}$
is an $\varepsilon$-approximation of $\sigma$.
In our algorithm, the resolutions are divided by half as the length level~$l_\ell$ and angle level~$l_\theta$ increase.
Thus, there exists a fine-enough resolution 
$\tilde{R} = \{2^{-l_\ell}\cdot\delta\ell_{\max}, 2^{-l_\theta}\cdot\delta\theta_{\max}\}$
that satisfies
$2^{-k_\ell}\cdot\delta\ell_{\max} < r_\ell$,
$2^{-k_\theta}\cdot\delta\theta_{\max} < r_\theta$.
Setting the cutoff resolution~$R_{\min}$ to be finer (both with respect to the insertion as well as rotation) than $\tilde{R}$
ensures that $M_{R(\sigma, \varepsilon)}$ can be approximated arbitrarily well.\footnote{To be more precise, one needs to account for the cases where $R(\sigma, \varepsilon_r)$ is not in the sequence of resolutions considered by the algorithm and we may introduce additional error when approximating $R(\sigma, \varepsilon_r)$ with $\tilde{R}$. 
However, using the techniques we previously used this can be easily accounted for. We omit this in our proof sketch. }

The search tree built with \rcsnr is a subtree of a dense tree in which each node is expanded with every element in 
$\M_{\rm fs}(\tilde{R}, \{0, \kappa_{\max}\})$.
This is because every coarse motion primitive used in \rcsnr can be decomposed into a sequence of motion primitives in $\M_{\rm fs}(\tilde{R}, \{0, \kappa_{\max}\})$.
Additionally, if we allow the algorithm run until the OPEN list is exhausted, every node in the dense tree (except for those that are in collision) will be explored by \rcsnr.
Since the dense tree encodes all possible trajectories that can be decomposed with $\M_{\rm fs}(\tilde{R}, \{0, \kappa_{\max}\})$,
when the solution $\sigma$ is traceable, has $\gamma$-clearance, and the system is Lipschitz continuous, an $\varepsilon$-approximation (with $\varepsilon < \gamma$) of $\sigma$ will be encoded in the dense tree and thus will be explored by \rcsnr.
$\hfill\blacksquare$

\subsection{Resolution completeness (with similar node rejection)}
\label{subsec:node-rejection}

We are now ready to show that even with similar node rejection, our algorithm is still resolution complete

\begin{thm}[Resolution completeness with similar-node rejection]
\label{thm:resolution_complete2}
    Let $\Delta = (\X, \W_{\rm obs}, \mathbf{x}_{\rm start}, p_{\rm goal}, \tau, \ell_{\max}, \kappa_{\max})$ be a steerable needle motion planning problem.
    \rcsb will find a solution in finite time, if the following conditions are satisfied:
    \begin{enumerate}[label=(\textbf{C\arabic*})]
        \item 
        \label{C1}
        The system is Lipschitz continuous.

        \item
        \label{C2}
        The cutoff resolution~$R_{\min}$ is fine enough and it satisfies 
        $\delta\ell_{\min} = 2^{-l_{\ell\max}}\cdot\delta\ell_{\max}, \delta\theta_{\min} = 2^{-l_{\theta\max}}\cdot\delta\theta_{\max}$.

        \item
        \label{C3}
        The radius~$d_{\rm sim}$ used to reject similar nodes satisfies 
        $$
            0 < d_{\rm sim} < {\min}
            \bigg\{
             \delta\ell_{\min}, \frac{\tau(L_s - 1)}{2(L_s^H - 1)}
            \bigg\},
        $$
        where $H = \lceil \ell_{\max}/\delta\ell_{\min}\rceil$.
        
        \item 
        \label{C4}
        There exists a traceable solution plan $\sigma$ 
        with $\frac{\tau}{2}$ goal tolerance and strong $\gamma$-clearance (Def.~\ref{dft:clearance}) for $\gamma > \frac{\tau + \delta\ell_{\min}}{2}$.
    \end{enumerate}
\end{thm}

The proof of Thm.~\ref{thm:resolution_complete2} uses Thm.~\ref{thm:resolution_complete1} and then follows Cheng and LaValle~\cite[Thm. 5.2]{Cheng2002_ICRA}. We include it here for completeness.

\vspace{2mm}
\textit{Proof.}
According to Thm.~\ref{thm:resolution_complete1}, \rcsnr terminates in finite time, thus \rcsb also terminates in finite time since more nodes are rejected.
We now prove that \rcsb can find a solution plan if conditions~\ref{C1}-\ref{C4}  are satisfied.

Since $\sigma$ is traceable, there exists some fine resolution~$R(\sigma, \varepsilon)$ can be explored by \rcsb (as discussed in Thm.~\ref{thm:resolution_complete1}), with which we can construct an $\varepsilon$-approximation of $\sigma$.
Denote the decomposable approximation $\sigma'$, and the sequence of motion primitives to compose it 
$M_{\sigma'} = \{\M_1,\dots,\M_n\}$.
When 
$M_{\sigma'}$ is sequentially applied to 
$\mathbf{x}_{\rm start}$, 
we obtain a sequence of configurations
$\{\mathbf{x}_0,\mathbf{x}_1,\dots,\mathbf{x}_n\}$, 
where $\mathbf{x}_0 = \mathbf{x}_{\rm start}, \mathbf{x}_i = \mathbf{x}_{i-1} \oplus \M_i, i \in [1, n]$.
For the rest of the proof, we use $M_{\sigma}[i,j] = \{\M_i, \dots, \M_j\}$ to denote a subsequence of $M_{\sigma'}$.
We also use $\mathbf{x} + M_{\sigma'}[i,j]$ to denote the configuration after sequentially applying $\{\M_i, \dots, \M_j\}$ to $\mathbf{x}$.

If we run \rcsnr, every $\mathbf{x}_i$ will be explored and $\sigma$ will be constructed when the search terminates.
However, if we run \rcsb, we prune nodes using duplicate detection (Sec.~\ref{subsec:duplicate_detection}). 
Thus, we need to show that even with pruning, \rcsb will still find a plan.
This will be done by showing that the same sequence of motion primitives can be applied to configurations that are ``similar'' to $\mathbf{x}_0 \ldots \mathbf{x}_n$ and the resultant plan $\tilde{\sigma}$ exists using the fact that $\tilde{\sigma}$ is ``similar'' to~$\sigma$ and that $\sigma$ has $\gamma$-clearance.
The rest of this proof formalizes this idea.

\begin{figure}
    \centering
    \includegraphics[width=\linewidth]{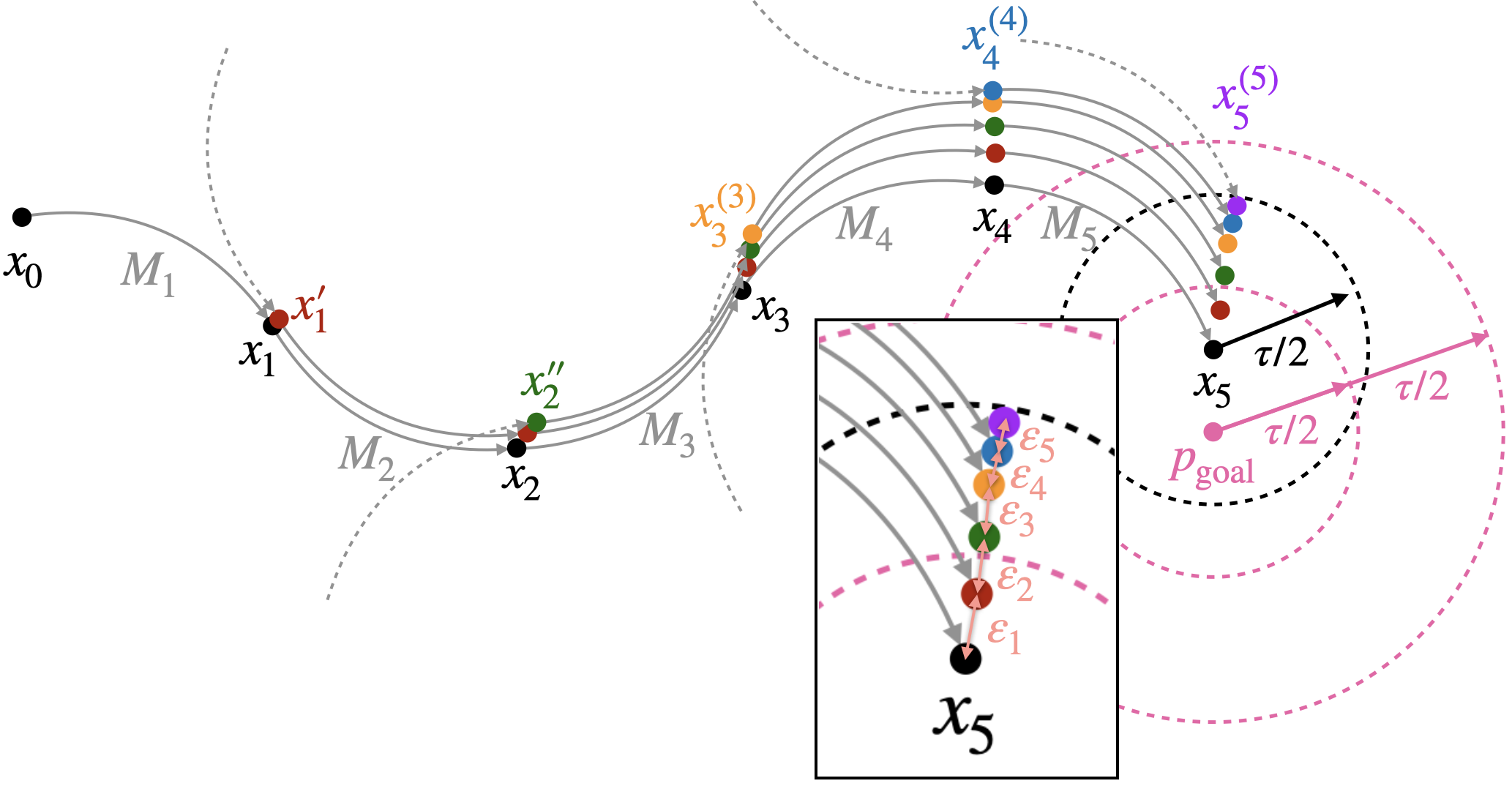}
    \caption{A 2D illustration of configuration pruning. 
    $\sigma$ is shown as black nodes, 
    the plan after~$\mathbf{x}'_1$ prunes~$\mathbf{x}_1$ is shown as red nodes,
    the plan after~$\mathbf{x}''_2$ prunes~$\mathbf{x}'_2$ is shown as green nodes,
    the plan after~$\mathbf{x}^{(3)}_3$ prunes~$\mathbf{x}''_3$ is shown as yellow nodes,
    the plan after~$\mathbf{x}^{(4)}_4$ prunes~$\mathbf{x}^{(3)}_4$ is shown as blue nodes,
    and the pruning configuration $\mathbf{x}^{(5)}_5$ is shown as a purple node.
    The solid circular arrows represent elements in $M_{\sigma}$, and the dashed circular arrows represent connections to predecessors of the pruning configurations.
    In this particular example, as long as we guarantee that $\|{\rm Proj}(\mathbf{x}_5) - p_{\rm goal}\|_2 \leq \frac{\tau}{2}$ and that~$\varepsilon = \sum_{i=1}^5\varepsilon_i \leq \frac{\tau}{2}$, the resultant plan which ended at $\mathbf{x}^{(5)}_5$ still satisfies the required goal tolerance.
    }
    \label{fig:completeness}
\end{figure}

Recall that~\ref{C3} ensures that $d_{\rm sim} < \delta\ell_{\min}$ which guarantees that any motion primitive will end up at a non-similar configuration.
Now, let $\mathbf{x}_i$ be the first configuration that is pruned because of a similar configuration (see Alg.~\ref{alg:main}, line~\ref{line:similar_node}).
We will say that $\mathbf{x}_i$ is \emph{replaced} by the similar configuration~$\mathbf{x}'_i$.
As~$i\geq 1$, in the worst case we have $i = 1$.
We then apply~$M_{\sigma'}[2,n]$ to $\mathbf{x}'_1$.
According to~\ref{C1}, the maximal error accumulated to $\mathbf{x}'_n = \mathbf{x}'_1 + M_{\sigma'}[2,n]$ is  
$\varepsilon_1 = \rho(\mathbf{x}'_n, \mathbf{x}_n) = L_s^{n - 1}\cdot d_{\rm sim}$.
Similarly, when $\mathbf{x}'_2$ is replaced by $\mathbf{x}''_2$, we apply $M_{\sigma'}[3,n]$ to $\mathbf{x}''_2$ and for $\mathbf{x}''_n = \mathbf{x}''_2 + M_{\sigma'}[3,n]$, the accumulated error is
$\varepsilon_2 = \rho(\mathbf{x}''_n, \mathbf{x}'_n) = L_s^{n - 2}\cdot d_{\rm res}$.
The same analysis applies for $\{\mathbf{x}_3,\dots,\mathbf{x}_n\}$.
According to~\ref{C3}, the total accumulated error then becomes:
\begin{equation*}
\begin{split}
    \varepsilon 
    &= \rho(\mathbf{x}^{(n)}_n, \mathbf{x}_n) 
    \leq \rho(\mathbf{x}'_n, \mathbf{x}_n) + \dots + \rho(\mathbf{x}^{(n)}_n, \mathbf{x}^{(n-1)}_n)   \\
    &= \varepsilon_1 + \dots + \varepsilon_n 
    = \frac{L_s^n - 1}{L_s -1}\cdot d_{\rm res} 
    < \frac{\tau}{2}\cdot \frac{L_s^n - 1}{L_s^H -1}
    \leq \frac{\tau}{2}.
\end{split}
\end{equation*}
According to~\ref{C4}, we have that $\|{\rm Prog}(\mathbf{x}_n) - g_{\rm goal}\|_2 \leq \frac{\tau}{2}$. Thus,
\begin{equation*}
\begin{split}
    &\|{\rm Proj}(\mathbf{x}^{(n)}_n) - p_{\rm goal}\|_2 \\
    &\leq \|{\rm Proj}(\mathbf{x}^{(n)}_n) - {\rm Proj}(\mathbf{x}_n)\|_2 + \|{\rm Proj}(\mathbf{x}_n) - p_{\rm goal}\|_2 \\
    &\leq \tau/2 + \tau/2 = \tau.
\end{split}
\end{equation*}
This implies that even in the worst case where all possible replacements happen, the final configuration $\mathbf{x}^{(n)}_n$ still satisfies the required goal tolerance (see Fig.~\ref{fig:completeness}).

Additionally, we prove that when pruning happens for~$\mathbf{x}^{(j)}_i$, the motion plan constructed with $M_{\sigma'}[i, n]$ is still collision-free.
We have shown above that $\rho(\mathbf{x}^{(n)}_n, \mathbf{x}_n) < \frac{\tau}{2}$. Moreover, we have that $\forall i \in [0, n], \rho(\mathbf{x}^{(n)}_i, \mathbf{x}_i) < \frac{\tau}{2}$ since less error is accumulated for $i < n$.
Thus we have that
$$
\forall k \in [i, n],
\|{\rm Proj}((\mathbf{x}^{(j)}_k) - {\rm Proj}((\mathbf{x}_k)\|_2 \leq \rho(\mathbf{x}^{(j)}_k, \mathbf{x}_k) < \frac{\tau}{2}.$$
And for any configuration $\tilde{\mathbf{x}}$ along edge $(\mathbf{x}^{(j)}_k, \mathbf{x}^{(j)}_{k+1})$, we have that
\begin{equation*}
\begin{split}
    {\min}\{ &  \|{\rm Proj}(\tilde{\mathbf{x}}) - {\rm Proj}(\mathbf{x}_k)\|_2 \\
    &\|{\rm Proj}(\tilde{\mathbf{x}}) - {\rm Proj}(\mathbf{x}_{k+1})\|_2\} < \frac{\tau + \delta\ell_{\min}}{2}.
\end{split}
\end{equation*}

According to~\ref{C4}, $\gamma > \frac{\tau + \delta\ell_{\min}}{2}$,
there always exist a small value 
$\varepsilon = \gamma - \frac{\tau + \delta\ell_{\min}}{2}$.
Thus, as long as $\sigma'$ is an $\varepsilon$-approximation of $\sigma$,
$\tilde{\sigma}$ is then guaranteed to be a $\gamma$-approximation of $\sigma$.
$\sigma$'s strong $\gamma$-clearance guarantees that the motion plan constructed with $M_{\sigma'}[i, n]$ is collision-free.

To summarize, as long as the required conditions are satisfied, \rcsb still finds a motion plan.
$\hfill\blacksquare$

\subsection{Resolution completeness while incorporating implementation details}
\label{subsec:implementation-details2}
\begin{cor}
\label{cor:optimization}
    \rcs and \rcspara will also find a solution in finite time, if the conditions in Thm.~\ref{thm:resolution_complete2} are satisfied.
    In other words, none of the implementation details hinder the resolution completeness guarantees.
\end{cor}

\vspace{2mm}
\textit{Proof sketch.}
For \rcs, 
goal reachability checks only reject invalid nodes, 
direct goal connection  only provides early terminations without affecting the search tree, 
and equivalent-node pruning provides an efficient way to reject identical configurations early.

For \rcspara,
parallelization may change the order of processing nodes, but does not change the essence of the proofs.
Thus, \rcs and \rcspara also find a motion plan as \rcsb does.
$\hfill\blacksquare$

\section{Planner parameters for evaluation}
\label{sec:appendix-planner-params}

In this section we describe the parameters used by each planner. 
For the precise definition of the different parameters, the reader is referred to the original papers describing the \rrt-based algorithm~\cite{Patil2014_TRO} and \aft~\cite{Pinzi2019_IJCARS}.
\begin{enumerate}[label=(\roman*)]
    \item \rrt: We set goal biasing:~$5\%$ and direct goal connecting ratio:~$100\%$. The multi-threaded version of \rrt, denoted as \rrtpara, was implemented with Motion Planning Templates  (MPT)~\cite{Ichnowski2019_ICRA} and used 60 threads.
    \item \aft: We used two tree refinements and used the cost function defined in Eq.~\ref{eq:cost}.
    Additionally, we used five levels of increments of the fractal structure, a tree density of $17$, and we rotated the tree around the root axis $10$ times, each time with a step of~$\frac{\pi}{5}{\rm rad}$ (these values were chosen according to the analysis provided by Pinzi et al.~\cite{Pinzi2019_IJCARS}).
    \item \rcs: 
    The system cutoff resolution is computed for control frequency $40{\rm Hz}$, which corresponds to a time interval of~$0.025 {\rm s}$:~$\delta\ell_{\min} = 5 ({\rm mm /s})\cdot 0.025 {\rm s}= 0.125{\rm mm}, \delta\theta_{\min} = 2\pi ({\rm rad/s})\cdot 0.025 {\rm s} \approx 0.157{\rm rad}$.
    The value of insertion and rotation velocities are taken from~\cite{Rucker2013_TRO} and the control frequency is the measurement rate of the NDI Aurora tracking system~\cite{NDI}.
    Maximum length step:~$\delta\ell_{\max} = 20{\rm mm}$.
    Distance metric weighting parameter:~$\alpha = 0.05$.
    In addition, we empirically determined that~$d_{\rm sim} = 5.5e-5$. 
    We use 60 threads for \rcspara, the multi-threaded version of \rcs.
\end{enumerate}

As is mentioned in Sec.~\ref{sec:pdef}, for \rcs, we determine the finest resolution by considering the hardware's ability to measurably change the steerable needle tip's position and orientation in tissue.
We use conventional constant insertion and rotational velocities (as are commonly used in steerable needle robots) and the magnetic tracker reading frequency (commonly used for tracking steerable needle tips) to determine the minimal motions.
These real-world minimal motions of the steerable needle tip are the minimal motions explored by the search.
\section{Additional experiments}
\label{sec:appendix-experiments}

In this section we present additional experiments evaluating the quality of the plans produced by each planner.
More specifically, we focus on the trajectory length~$\ell(\sigma)$ and the final targeting error~$\|\sigma(1) - p_{\rm goal}\|_2$.

For \aft, both are considered in the cost function.
For \rcs and \rrt, although the plan quality is not explicitly optimized, as more running time is given, there is a chance to improve the plan quality.
For both planners, we use the same cost function defined in Eq.~\ref{eq:cost} to pick a plan with the lowest cost from all motion plans generated.
Since \rcs and \rrt only consider a plan to be valid if it satisfies the required targeting error, the final resulting plan is guaranteed to satisfy the targeting error.
Similar to the previous comparison, \rrtpara and \rcspara are allotted a running time of $100$ seconds,
and the planner keeps running after the first solution is found to generate more plans.
We pick as the final result the solution with minimal cost among all solutions found.
\aft uses two tree refinements.
The results are shown in Table~\ref{tab:comparison}.
Since different test cases have different ranges of plan length, we take the best plan produced by \rcspara as the baseline, and compute the plan length relative to it.
Values in Table~\ref{tab:comparison} are averaged over all test cases that are successfully solved by all planners.

\begin{table}[!t]
\caption{Planner Performance Comparison}
\label{tab:comparison}
\begin{center}
\begin{tabular}{|c||c||c||c|}
\hline
 & \rrtpara & \aft & \ \rcspara\\
\hline
Success rate & $91.2\%$ & $65.8\%$ & $97.6\%$ \\
\hline
Avg. relative length & $0.998$ & $1.003$ & $1.0$ \\
\hline
Avg. targeting error & $0.053 {\rm mm}$ & $0.207 {\rm mm}$ & $0.051 {\rm mm}$ \\
\hline
\end{tabular}
\end{center}

\vspace{-5mm}
\end{table}

Due the limited insertion length and the maximum curvature constraint, all three planners produced plans with (roughly) similar lengths.
\rrtpara computed the lowest-cost trajectory on average.
This may due to the steer function in \rrt always trying to connect to a sampled point with the shortest arc.
For the two search-based methods, \rcspara achieved better plan length since the resolution in \rcs can be much finer than that of \aft.
As for the targeting error, because \rcspara and \rrtpara both try to connect to the goal point directly, they can efficiently reduce the targeting error and achieve average targeting errors much smaller than~\aft.

\end{document}